\documentclass{article}
\usepackage{graphicx, epsfig, comment}
\usepackage{times,avant}
\usepackage{amsmath,amssymb,amsthm}
\usepackage{natbib}
\usepackage{subfigure}
\usepackage{algorithm}
\usepackage{algorithmic}
\usepackage{slashbox}
\usepackage[dvipsnames,usenames]{color}
\usepackage{multirow}
\def\RR{\mathbb{R}}

\def\cov{{\rm Cov}}

\def\set#1{\left\{#1\right\}} % a set
\def\bE{\mathbb E}

\def\cP{{\mathcal P}}
\def\cS{{\mathcal S}}

\def\SS{\mathbb{S}}
\def\RR{\mathbb{R}}

\def\cov{{\rm Cov}}

\def\loc{\tiny \mbox{loc}}
\DeclareMathOperator*{\argmin}{arg\,min}

\newcommand{\iid}{\stackrel{\mathrm{iid}}{\sim}}
\newcommand{\eqsvd}{\stackrel{\mathrm{svd}}{=}}

\newcommand{\vecthree}[3]
{
\begin{pmatrix} #1 \\ #2 \\ #3 \end{pmatrix}
}
\newcommand{\vectwo}[2]
{
\begin{pmatrix} #1 \\ #2 \end{pmatrix}
}
\newcommand{\mattwo}[4]
{
\begin{pmatrix} #1 & #2\\ #3 & #4\end{pmatrix}
}
\newcommand{\diagthree}[3]
{
\begin{pmatrix} #1 & 0 & 0\\ 0 & #2 & 0 \\ 0 & 0 & #3\end{pmatrix}
}

\theoremstyle{definition}

\newcounter{ex}
\begin{document}
\title{Randomized Dimension Reduction on Massive Data}
\author{Stoyan Georgiev and Sayan Mukherjee\footnote{Stoyan Georgiev is a Postdoctoral
Scholar in the Genetics Department at Stanford University, Palo Alto, CA 94305, U.S.A.
(email:sgeorg@stanford.edu). Sayan Mukherjee is an Associate Professor in the Departments of Statistical Science, Computer
Science and Mathematics, and the Institute for Genome Sciences \& Policy at Duke University, Durham, NC
27708-0251, U.S.A. (email:sayan@stat.duke.edu).}}
\date{\today}
\maketitle
\abstract{\noindent Scalability of statistical estimators is of increasing importance in 
modern applications and dimension reduction is often used to extract relevant information 
from data. A variety of popular dimension reduction approaches can be framed as symmetric 
generalized eigendecomposition problems. In this paper we outline how taking into account 
the low rank structure assumption implicit in these dimension reduction approaches provides 
both computational and statistical advantages. We adapt recent randomized low-rank 
approximation algorithms to provide efficient solutions to three dimension reduction 
methods: Principal Component Analysis (PCA), Sliced Inverse Regression (SIR), and 
Localized Sliced Inverse Regression (LSIR). A key observation in this paper is that 
 randomization serves a dual role, improving both computational and statistical 
performance. This point is highlighted in our experiments on real and simulated data.}
\bigskip

\noindent {\bf Key Words}: dimension reduction, generalized eigendecompositon, low-rank, 
 supervised, inverse regression, random projections, randomized algorithms, Krylov subspace methods
\newpage

%%1. sensitivity to Lncozs approximation perentage
%%2. actual rsvd algorithm used -- add to the appendix matrinsson's algorithm
%3. make explicit the application of RSVD in the context of SIR/LSIR: adaptive estimation of the
%of rank of $\Gamma$ -- r. The number of edrs is assumed to be given, if not report the 
%top r directions ordered by the estimates of the generalized eigenvalues.
%The number of 'stabl'e edrs can be estimated using  RSVD applied to 
%$\Sigma ^{-1/2}\Gamma$.
%%4. make CRAN package radr -- Randomized Adaptive Dimension Reduction  
%%5. put on Arxiv and submit to JCGS

\section{Introduction}
In the current era of information, large amounts of complex high dimensional
data are routinely generated across science and engineering. Estimating and
understanding the underlying structure in the data and using it to model scientific problems
is of fundamental importance in a variety of applications. As the size of the data sets
increases, the problem of statistical inference and computational feasibility become
inextricably linked. Dimension reduction is a natural approach to
summarizing massive data and has historically played a central role in data analysis,
visualization, and predictive modeling. It has had significant impact on both the
statistical inference
\citep{Adcock1878,Edgeworth1884,Fisher1922,Hotelling1933,Young1941}, as well as on
the numerical analysis research and applications
\citep{Golub:1969,golub:van_loan,Gu_Eisenstat_1996,Golub00:gsvd},
for a recent review see \citep{Mahoney:review}.
Historically, statisticians have focused on studying the theoretical properties
of estimators. Numerical analysts and computational mathematicians,
on the other hand, have been instrumental in the development of powerful
algorithms, with provable stability and convergence guarantees.
Naturally, many of these have been successfully
applied to compute estimators grounded on solid statistical foundations.
A classic example of this interplay is Principal Components Analysis (PCA)
\citep{Hotelling1933}. In PCA a target estimator is defined based on statistical
considerations about the sample variance in the data, which can then be efficiently
computed using a variety of Singular Value Decomposition (SVD) algorithms
developed by the numerical analysis community.

In this paper we focus on the problem of dimension reduction and
integrate the statistical considerations of \textit{estimation accuracy}
and out-of-sample errors with the numerical considerations of runtime
and \textit{numerical accuracy}.
Our proposed methodology builds on a classical approach to modeling large data
which first compresses the data, with minimal loss of relevant information,
and then applies statistical estimators appropriate for small-scale problems.
In particular, we focus on dimension reduction via (generalized) eigendecompositon
as the means for data compression and out-of-sample residual errors
as the measure of information.
%This idea underlies a variety of approaches
%to linear and non-linear dimension reduction both in the unsupervised and the
%supervised setting.
The scope of the current work encompasses many dimension reduction methods
which are implemented as solutions to truncated generalized eigendecomposition problems
\citep{Hotelling1933, LDA, Li:1991, lsir:2010}.%% , misha1, donoho03hessian, lle}.
We use randomized algorithms, developed in the
numerical analysis community \citep{Drineas:Ravi:Mahoney:2006,
Sarlos:random_projections:2006,Liberty:Woolfe,
Boutsidis:Drineas:2009, Rokhlin2008, Tropp2010}, to simultaneously
reduce the dimension and the impact of
independent random noise. 
%In Section \ref{sec:methods} we start by briefly introducing of the 
%relevant statistical and computational considerations of randomized
%algorithms for approximate low-rank approximation. Then in 
%Section \ref{sec:rsvd}  we propose an adaptive algorithm for approximate
%singular value decomposition (SVD) in which both the number of
%singular vectors as well as the number of numerical
%iterations are inferred from the data, based on statistical criteria.
%In Section \ref{sec:gegein} we focus on the  truncated generalized
%eigendecomposition problem and propose estimators for two related supervised 
%dimension reduction approaches SIR \citep{Li:1991}, and LSIR \citep{lsir:2010}.
%In Section \ref{sec:results}
%we demonstrate on simulated and real data examples that the randomized
%estimators provide not only a computationally attractive solution, but also
%in some cases improved statistical accuracy. We argue this is due to implicit
%regularization imposed by the randomized approximation.
There are three key ideas we develop in this paper:
\begin{enumerate}
\item[(1)] We propose an adaptive algorithm for approximate
singular value decomposition (SVD) in which both the number of
singular vectors as well as the number of numerical
iterations are inferred from the data, based on statistical criteria.
\item[(2)] We use the adaptive SVD algorithm to construct truncated generalized
eigendecomposition estimators for two supervised dimension reduction
approaches \citep{Li:1991, lsir:2010}.
\item[(3)] We demonstrate on simulated and real data examples that the randomized
estimators provide not only a computationally attractive solution, but also
in some cases improved statistical accuracy. We argue this is due to implicit
regularization imposed by the randomized approximation.
\end{enumerate}
In Section \ref{sec:methods} we describe the adaptive randomized
SVD procedure we use for the various dimension reduction methods.
In Section \ref{sec:gegein} we provide randomized estimators for
sliced inverse regression (SIR) \cite{Li:1991} and localized sliced
inverse regression (LSIR) \cite{lsir:2010}. Finally,
in Section \ref{sec:results} we illustrate the utility of the proposed
methodology on simulated and real data.

\section{Randomized algorithms for dimension reduction}
\label{sec:methods}
In this section %, after a brief summary of some key ideas underlying
%dimension reduction in the presence of noise and randomized matrix factorizations,
we develop algorithmic extensions to three dimension reduction
methods: PCA, SIR, and LSIR.
The first algorithm
provides a numerically efficient and statistically robust estimate of the highest
variance directions in the data using a randomized algorithm for singular
value decomposition (Randomized SVD) \citep{Rokhlin2008, Tropp2010}.
In this problem the objective is linear unsupervised dimension reduction with
the low-dimensional subspace estimated via an eigendecomposition.
Randomized SVD will serve as the core computational engine for the other two
dimension reduction estimators in which estimation reduces to a truncated
generalized eigendecomposition problem. The second algorithm
computes a low dimensional linear subspace that captures the
predictive information in the data. This is a supervised setting, in
which the input data consist of a set of features and a univariate response
for each observation. We focus on SIR \citep{Li:1991}
as it applies to both categorical and continuous responses and
subsumes the widely used Linear Discriminants Analysis (LDA)
\citep{LDA} as a special case. The third method to which we apply
randomization ideas is LSIR \citep{lsir:2010}, which provides linear
reduction that capture non-linear structure in the data using localization.
In the Appendix, we outline extensions of the developed
ideas to unsupervised manifold learning
\citep{isomap,lle,donoho03hessian,misha1}, with specific focus on
locality preserving projections (LPP) \citep{LPP:NIPS:2003}.
%For all methods we will also provide adaptive algorithms to estimate
%the dimension of the projective linear subspace.
%In the next section we outline the main computational and
%statistical considerations underlying our proposed methodology.

\subsection{Notation}
Given positive integers $p$ and $d$ with $p \gg d$, $\RR^{p \times d}$ stands
for the class of all matrices with real entries of dimension $p \times d$,
and $\SS^{p}_{++}$ ($\SS^{p}_{+}$) denotes the sub-class of
symmetric positive definite (semi-definite)
$p \times p $ matrices. For $B \in \RR^{p \times d}$, span($B$) denotes the
subspace of $\RR^p$ spanned by the columns of $B$. A \textit{basis matrix}
for a subspace $\cS$ is any full column rank matrix $B \in \RR^{p \times d}$
such that $\cS=\text{span}(B)$, where $d=\text{dim}(\cS)$. In the case of sample
data $X$, eigen-basis($X$) denotes the orthonormal left eigenvector basis.
Denote the data matrix by $X=(x_1,\ldots,x_n)^T \in \RR^{n \times p}$, where each
sample $x_i$ is a assumed to be generated by a $p$-dimensional probability
distribution $\cP_{_X}$. In the case of supervised dimensions reduction, denote the
response vector to be $Y \in \RR^{n}$ (quantitative response) or $Y \in \{1,\ldots,C\}$
(categorical response with $C$ categories), and $Y \sim \cP_{_Y}$. The data and
the response are assumed to have a joint distribution $(X, Y) \sim \cP_{_{X \times Y}}$.
Unless explicitly specified otherwise, assume that both the sample data and the response 
(for the regression case) are centered independently for the \textit{training} and the 
\textit{test} data. Hence, $\sum_{i=1}^n y_i = \sum_{i=1}^n x_{ij} = 0 \ \text{for all} \ j=1,\ldots,p$.

\subsection{Computational considerations}
\label{sec2:computational}
The main computational tool used in our development is a randomized
algorithm for approximate eigendecompositon, which
factorizes a $n \times p$ matrix of rank $r$ in time
${\cal O}(npr)$ rather than the ${\cal O}(np \times \mbox{min}(n,p))$ required
by approaches that do not take advantage of the special structure
in the input. This is particularly relevant to statistical applications
in which the data is high dimensional but reflects a highly constrained process
e.g. from biology or finance applications, which suggests it has low
intrinsic dimensionality i.e. $r \ll n < p$. %Given the low rank factorization,
%matrix multiplications in the low dimensional space replace those in the
%ambient space with little loss in accuracy of the matrix operations.
An appealing characteristic of the employed randomized algorithm
is the explicit control of the tradeoff between estimation accuracy
relative to the exact sample estimates and computational efficiency.
Rapid convergence to the exact sample estimates can be achieved
and was investigated in \citep{Rokhlin2008}.

\subsection{Statistical considerations}
\label{sec2:stats}
A central concept in this paper is that the randomized approximation
algorithms we use for statistical inference impose regularization
constraints.
Thinking of the estimate computed by the randomized
algorithm as a statistical model helps highlight this idea.
%Consider two estimators, one corresponding to the exact (generalized)
%eigendecomposition and the other corresponding to the randomized
%algorithm.
The numerical analysis perspective typically assumes a deterministic
view of the data input, focusing on the discrepancy between the
randomized and the exact solution which is estimated and evaluated
on the same data.
%The goal is to minimize the difference
%between the estimates, irrespective of the data generating process.
However, in many practical applications the data is a noisy random
sample from a population. Hence, when the
interest is in dimension reduction, the relevant error comparison
is between the true subspace that captures information about
the population and the corresponding algorithmic estimates. The
true subspace is typically unknown which makes it necessary to use
proxies such as estimates of the out-of-sample generalization
performance.
A key parameter of the randomized estimators, described in detail
in Section \ref{sec:rsvd}, is the number of power iterations used to
estimate the span of the data. Increasing values for that parameter
provide intermediate solutions to the factorization problem which
converge to the exact answer \citep{Rokhlin2008}. Fewer iterations
correspond to reduced runtime but also to larger deviation from
the sample estimates and hence stronger regularization.
We show in Section \ref{sec:results} the regularization effect of
randomization on both simulated as well as real data sets and
argue that in some cases fewer iterations may be justified
not only by computational, but also by statistical considerations.

\subsection{Adaptive randomized low-rank approximation}\label{sec:rsvd}
In this section we provide a brief description of a randomized estimator for the best low-rank matrix
approximation, introduced by \citep{Rokhlin2008, Tropp2010}, which combines random projections
with numerically stable matrix factorization. We consider this
numerical framework as implementing a computationally efficient
shrinkage estimator for the subspace capturing the largest variance
directions in the data, particularly appropriate when applied to input
matrix of (approximately) low rank. Detailed discussion of the estimation accuracy of Randomized SVD
in the absence of noise is provided in \citep{Rokhlin2008}.
The idea of random projection was first developed as a proof
technique to study the distortion induced by low dimensional
embedding of high-dimensional vectors in the work of
\citep{Johnson1984} with much literature simplifying and sharpening
the results \citep{JLS:Frankl:1987, Indyk:1998,JLS:simplified_proof:1999, Achlioptas:2001}.
More recently, the theoretical computer science and the numerical analysis
communities discovered that random projections can be used for
efficient approximation algorithms for a variety of applications
\citep{Drineas:Ravi:Mahoney:2006, Sarlos:random_projections:2006, Liberty:Woolfe,
Boutsidis:Drineas:2009, Rokhlin2008, Tropp2010}.
We focus on one such approach proposed by \citep{Rokhlin2008, Tropp2010} which
targets the accurate low-rank approximation of a given large data matrix $X \in \RR^{n \times p}$.
In particular, we extend the randomization methodology to the noise setting, in which the
estimation error is due to both the approximation of the low-rank structure in $X$,
as well as the added noise. A simple working model capturing this scenario is
as follows: $X = X_{d^{*}} + E$, where $X_{d^{*}} \in \RR^{n \times p}$, rank$(X_{d^{*}})=d^{*}$
captures the ``signal", while $E$ is independent additive noise.
\paragraph{Algorithm.}
Given an upper bound on the target rank $d_{\max}$ and on the number of necessary
power iterations $t_{\max}$ ($t_{\max} \in \{5,\ldots 10\}$ would be sufficient in most cases),
the algorithm proceeds in two stages: (1) estimate a basis for the range
of $X_{d^{*}}$, (2) project the data onto this basis and apply SVD:
\centerline{Algorithm: \textit{Adaptive Randomized SVD}(X, $t_{\max}$, $d_{\max}$, $\Delta$) }\label{alg:arsvd}
\begin{enumerate}
\item[(1)] Find orthonormal basis for the range of $X$;
\begin{enumerate}
\item[(i)] Set the number working directions: $\ell = d_{\max} + \Delta$;
\item[(ii)] Generate random matrix: $\Omega \in \RR^{n \times \ell}$ with $\Omega_{ij} \stackrel{iid}{\sim} \mbox{N}(0,1)$;
\item[(iii)] Construct blocks: $F^{(t)} = X X^T F^{(t-1)}$ with $F^{(0)} = \Omega$ for $t \in \{1,\ldots,t_{\max}\}$;
\item[(iv)] Select optimal block $t^{*} \in \{1,\ldots,t_{\max}\}$ and rank estimate $d^{*} \in \{1,\ldots, d_{\max}\}$,
using a stability criterion and Bi-Cross-Validation (see Section \ref{rankest_unsup});
\item[(v)] Factorize selected block: $\ F^{(t^{*})} = Q S \in \RR^{n \times l},\ \ \ \ \ Q^{T}Q=I$;
\end{enumerate}
\item[(2)] Project data onto the range basis and compute SVD;
\begin{enumerate}
\item[(i)] Project onto the basis: $B = X^{T} Q \in \RR^{p \times \ell} $;
\item[(ii)] Factorize: $B \eqsvd U \Sigma W^{T}$, where $\Sigma = \text{diag(}\sigma_1,\ldots ,\sigma_{\ell}\text{)}$;
\item[(iii)] Rank $d^{*}$ approximation: $\widehat{X}_{d^{*}} = U_{d^{*}} \Sigma_{d^{*}} V_{d^{*}}^{T}$\\
$U_{d^{*}}= (U_1 | \ldots | U_{d^{*}} ) \in \RR^{n \times d^{*}} $\\
$\Sigma_{d^{*}}= \text{diag(}\sigma_1,\ldots ,\sigma_{d^{*}}\text{)} \in \RR^{d^{*} \times d^{*}} $\\
$V_{d^{*}} = Q \times (W_1|\ldots | W_{d^{*}}) \in \RR^{p \times d^{*}} $;
\end{enumerate}
\end{enumerate}

In stage (1) we set the number of working directions $\ell=d_{\max} + \Delta$ to be the sum of the upper bound
on the rank of the data, $d_{\max}$, and a small oversampling parameter $\Delta$, which ensures more stable
approximation of the top $d_{\max}$ sample variance direction (the estimator tends to be robust to changes in
$\Delta$, so we use $\Delta=10$ as a suggested default).
In step (iii) the random projection matrix $\Omega$ is applied to powers of $X X^T$ to
randomly sample linear combinations of eigenvectors of the data weighted by powers
of the eigenvalues
$$\underbrace{F^{(t)}}_{n \times \ell} = (XX^T)^{t}\Omega = US^{2t}U^T\Omega = US^{2t} \Omega^*, \quad \mbox{ where } X \eqsvd U S V^T.$$
The main goal of the power iterations is to increase the decay of
the noise portion of the eigen-spectrum while leaving the eigenvectors
unchanged. This is a regularization or shrinkage
constraint. Note that each column $j$ of $F^{(t)}$ corresponds
to a draw from a $n$-dimensional Gaussian distribution:
$F^{(t)}_j \sim N(0, U S^{4t} U^T)$, with the covariance structure
more strongly biased towards higher directions of variation as
$t$ increases. This type of regularization is analogous to the
local shrinkage term developed in \citep{ScottPolson}. In step (iv)
we select an optimal block $F^{(t^{*})}$ for $t^{*} \in \{1,\ldots,t_{\max}\}$
and estimate an orthonormal basis for the column space.
In \citep{Rokhlin2008} the authors assume fixed target rank $d^{*}$
and aim to approximate $X$, rather than $X_{d^{*}}$.
They show that the optimal strategy in that case is to set $t^{*}=t_{\max}$, which typically
achieves excellent $d^{*}$-rank approximation accuracy for $X$,
even for relatively small values of $t_{\max}$.
%but, as suggested by \citep{Rokhlin2008}, we prefer to use the entire collection
%$(F^{(1)}\ |\ F^{(2)}\ | \ldots|\ F^{(t)})$ to provide
%numerical stability in same spirit as the (blocked) Lanczcos approach (Chapter 9,
%in \citep{golub:van_loan}).
In this work we focus on the ``noisy" case, where $E\ne 0$ and propose
to adaptively set both $d^{*}$ and $t^{*}$ aiming to optimize
the generalization performance of the Randomized estimator.
The estimation strategy for $d^{*}$ and $t^{*}$ is described in detail
in Section \ref{rankest_unsup}.

In stage (2) we rotate the orthogonal
basis $Q$ computed in stage (1) to the canonical eigenvector
basis and scale according to the corresponding eigenvalues.
In step (i) the data is projected onto the low dimensional
orthogonal basis $Q$. Step (ii) computes exact SVD in the
projected space.

\paragraph{Computational complexity.}
The computational complexity of the randomization step is 
${\cal O}(np \times d_{\max} \times t_{\max})$ and the factorizations in the 
lower dimensional space have complexity
${\cal O}(n p \times d_{\max} + n \times d^2_{\max} )$.
With $d_{\max}$ small relative to $n$ and $p$,
% the resulting
%overall complexity, without taking into account the
%estimation of $d^{*}$ and $t^{*}$, is $O(t_{\max} d_{max} n p)$.
the runtime in both steps is dominated by the multiplication by
the data matrix and in the case of sparse data fast multiplication
can further reduce the runtime. We use a normalized version
of the above algorithm that has the same runtime complexity
but is numerically more stable \citep{normalized_rsvd:2010}.

\subsubsection{Adaptive method to estimate $d^{*}$ and $t^{*}$}\label{rankest_unsup}
We propose to use ideas of stability 
under random projections in combination with cross-validation
to estimate the intrinsic dimensionality of the dimension reduction
subspace $d^{*}$ as well as the optimal value of the eigenvalue 
shrinkage parameter $t^{*}$.

\paragraph{Stability-based estimation of $d^{*}$}
First, we assume the regularization parameter $t$ is fixed and describe 
the estimation of the rank parameter $d^{*}(t)$, using a stability
criterion based on random projections of the data.
We start with an upper-bound guess -- $d_{\max}$ for $d^{*}$ and apply a small number -- $B$
(e.g $B$ = 5) independent Gaussian random projections
$\Omega^{(b)} \in \RR^{n \times d_{\max}}$, $\Omega^{(b)}_{ij} \stackrel{iid}{\sim} \mbox{N}(0,1)$,
for $b \in \{1,\ldots,B\}$. Then we find an estimate of the
eigenvector basis of the column space of the projected data. Assuming approximately
low-rank structure for the data, we reduce the influence of the noise relative to
the signal we enhance the higher relative to the lower variance directions by
raising all eigenvalues to the power $t$:
$$
U_{b}^{(t)} \equiv (U_{b1}^{(t)} | \ldots | U_{bd}^{(t)} ) = \mbox{eigenbasis}[(XX^T)^t\Omega^{(b)}] \ \ \mbox{for} \ \ b \in \{1,\ldots,B\}.
$$
The $k$-th principal basis eigenvector estimate ($k \in \{1,\ldots,d\}$) is assigned a \textit{stability score}:
$$
\mbox{stab}(t,k,B) = \frac{1}{N} \sum_{j_1=1}^{B-1}\sum_{j_2=j_1+1}^{B}\left|\mbox{cor}(U_{j_1k}^{(t)}, U_{j_2k}^{(t)})\right|,
\ \ \ \mbox{where} \ N=\frac{B(B-1)}{2}.
$$
Here $U_{rk}^{(t)}$ is the estimate of the $k$-th principal eigenvector
of $X^TX$ based on the $r$-th random projection and
$\mbox{cor}(U_{j_1k}^{(t)}, U_{j_2k}^{(t)})$ denotes the
Spearman rank-sum correlation between $U_{j_1k}^{(t)}$ and
$U_{j_2k}^{(t)}$.
Eigenvector directions that are not dominated by independent noise are
expected to have higher stability scores. When the data has approximately
low-rank we expect a sharp transition in the eigenvector stability between
the ``signal" and ``noise" directions. In order to estimate this change point
we apply a non-parametric location shift test (Wilcoxon rank-sum) to each
of the $d_{\max} - 2$ stability score partitions of eigenvectors with larger vs.
smaller eigenvalues. The subset of principal eigenvectors that can be
stably estimated from the data for the given value of $t$ is determined
by the change point with smallest p-value among all $d_{\max} - 2$
non-parametric tests.
$$
\hat{d}_{t} = \argmin_{k \in\{2,\ldots,d_{\max} - 1\}} \mbox{p-value}(k,t)
$$
where $\mbox{p-value(k,t)}$ is the p-value from the Wilcoxon rank-sum test
applied to the $\{\mbox{stab}(t,i,B)\}_{i=1}^{k-1}$ and
$\{\mbox{stab}(t,i,B)\}_{i=k}^{d_{\max}}$.

\paragraph{Estimation of $t^{*}$}\label{sec:adapt_t}
%Following the stability-based estimation of $d^*(t)$, 
In this section we describe a procedure for selecting optimal value for  
$t \in \{1, \ldots, t_{\max}\}$ based on the reconstruction accuracy under 
Bi-Cross-Validation for SVD \citep{Owen2009}, using generalized Gabriel 
holdout pattern \citep{Gabriel2002}. The rows and columns of the input matrix
are randomly partitioned into $r$ and $c$ groups respectively, resulting in a total of
$r \times c$ sub-matrices with non-overlapping entries. %For each $t \in \{1, \ldots, t_{\max}\}$ 
We apply \textit{Adaptive Randomized SVD} to factorize the training data 
from each combination of $(r-1)$ row and $(c - 1)$ column blocks.
%containing the rows and columns from $(r-1) \times (c - 1)$ bocks. 
In each case the submatrix block with the omitted rows and columns is
approximated using its modified Schur complement in the training data. 
The cross-validation error for each data split corresponds to the Frobenius norm of the
difference between the held-out submatrix and its training-data-based estimate.
For each sub-matrix approximation we estimate the rank
$d^*= d(t)$ using the stability-based approach from the previous section.
As suggested in \citep{Owen2009}, we fix $c=r=2$, in which case
Bi-Cross-Validation error corresponding to holding out the top left block
$A$ of a given block-partitioned matrix $\mattwo{A}{B}{C}{D}$, becomes
$||A - BD^{+}C||^2_F$. Here $D^{+}=VS^{-1}U^T$ is the Moore-Penrose
pseudoinverse of $D\eqsvd USV^T$.
For fixed value of $t$ we estimate $d(t)$ and factorize $D^{+}$ using
\textit{Adaptive Randomized SVD(t, d(t), $\Delta=10$)} and denote
the Bi-Cross-Validation error by $\text{BiCV}(t, A)$. The same process
is repeated for the other three blocks B, C, and D. The final error and
rank estimates are defined to be the medians across all blocks and are
denoted as $\text{BiCV}(t)$ and $\hat{d}_t$, respectively.
We optimize over the full range of allowable values for $t$ to arrive at the final estimates
\begin{eqnarray*}
\hat{t}_{\text{BiCV}}^* &=& \argmin_{t \in \{1,\ldots,t_{\max} \}} \text{BiCV}(t)\\
\hat{d}_{\text{BiCV}}^* &=& \hat{d}_{\hat{t}_{\text{BiCV}}^*}.
\end{eqnarray*}

\subsection{Generalized eigendecomposition and dimension reduction}\label{sec:gegein}
In this section we discuss a formulation of the truncated generalized
eigendecomposition problem, particularly relevant to dimension reduction.
Our proposed solution is based on the \textit{Adaptive Randomized SVD}
from Section \ref{sec:rsvd} and is motivated by the estimation of the dimension
reduction subspace for SIR and LSIR.  In the Appendix we have 
outlined an application to nonlinear dimension reduction \citep{misha1,LPP:NIPS:2003}.

\subsubsection{Problem Formulation.} Assume we are given 
$\Sigma \in \SS^{p}_{++}, \Gamma \in \SS^{p}_{+}$ that
characterize pairwise relationships in the data and let $r \ll \min(n, p)$
be the "intrinsic dimensionality" of the information contained in the data.
In the case of supervised dimension reduction methods this corresponds to
the dimensionality of the linear subspace to which the joint distribution of
$(X,Y)$ assigns non-zero probability mass. Our objective is to find a basis
for that subspace. For SIR and LSIR this corresponds to the span of the
generalized eigenvectors $\{g_1,\ldots,g_{r}\}$ with largest eigenvalues 
$\{\lambda_{\max} = \lambda_1 \le \ldots \le\lambda_{r}\}$:
\begin{eqnarray}\label{eqn:gsvd:exact1}
\Gamma g=\lambda \Sigma g.
\end{eqnarray}
An important structural constraint we impose on $\Gamma$, which is
applicable to a variety of high-dimensional data settings, is that it has 
low-rank: $r \le d^{*} \equiv \mbox{rank}(\Gamma) \ll p$. 
It is this constraint that we will take advantage of in the 
randomized methods. In the case of $\Sigma = \mbox{I}$ (unsupervised case)
$r=d^{*}$.
%We assume $r$ is given and adaptively estimate $\mbox{rank}(\Gamma)$ 
%using the randomized algorithms from Section \ref{sec:rsvd}.

\subsubsection{Sufficient dimension reduction}\label{sec:sdr}
Dimension reduction is often a first step in the statistical analysis of
high-dimensional data and could be followed by data visualization or
predictive modeling. If the ultimate goal is the latter, then the statistical
quantity of interest is a low dimensional summary $Z \equiv R(X)$ which
captures all the predictive information in $X$ relevant to $Y$:
$$ Y = f(X) + \varepsilon = h(Z) + \varepsilon, \quad X\in \RR^{p}, Z \in \RR^{r}, \ r \ll p.$$
Sufficient dimension reduction (SDR) is one popular approach for estimating $Z$, which
%%\citep{Li:1991,Cook:Weis:1991,Li:1992,HT96:locLDA,nca,GL05,Li2005,Nilsson07,Sugiyama2007,Cook2007,lsir:2010}.
\citep{Li:1991, Cook:Weis:1991, Li:1992, Li2005,Nilsson07, Sugiyama2007, Cook2007, lsir:2010}.
%A very useful consequence for predictive
%modeling is the identity $\bE[Y \mid X] = \bE[Y \mid R(X)]$, see \citep{Cook2007}
%for more details.
In this paper we focus on linear SDRs:
$G=(g_{1},\ldots,g_{r}) \in \RR^{p \times r} \Rightarrow R(X) = G^T X$, which
provide a prediction-optimal reduction of $X$
$$(Y \mid X) \stackrel{d}{=} (Y \mid G^T X), \quad \stackrel{d}{=}
\mbox{ is equivalence in distribution}.$$
We will consider two specific supervised dimension reduction methods:
Sliced Inverse Regression (SIR) \citep{Li:1991} and Localized Sliced
Inverse Regression (LSIR) \citep{lsir:2010}. SIR is effective
when the predictive structure in the data is global, i.e. there is
single predictive subspace over the support of the marginal distribution of
$X$. In the case of local or manifold predictive structure in
the data, LSIR can be used to compute a projection matrix $G$ that contains this
non-linear (manifold) structure.

\paragraph{Inverse Regression (SIR)}\label{sec:sir}
Sliced Inverse Regression (SIR) is a dimension reduction approach
introduced by \citep{Li:1991}. The relevant statistical quantities
for SIR are the covariance matrix, $\Sigma = \cov(X)$, and
the covariance of the inverse regression, $\Gamma =
\cov[\bE_X(X|Y)]$. We use the formulation from (\ref{eqn:gsvd:exact1})
to find the dimension reduction subspace
$\hat{G} = \{\hat{g}_1,\ldots,\hat{g}_{r}\}$, based on observations 
$\set{(x_i, y_i)}_{i=1}^n$ and the empirical estimates
$\hat{\Sigma}$ and $\hat{\Gamma}$ (e.g. see \citep{Li:1991}).
If there exists a linear projection matrix $G$ with the property that
\begin{equation}
\label{sir:asm}
\mbox{span}(\bE_X[X \mid Y]-\bE_X[X]) \in \mbox{span}(\Sigma G)
\end{equation}
then SIR is effective. This assumption can problematic due to non-linear
predictive structure in the data e.g. manifold or clustering structure associated with
differences in the response. SIR has been adapted to address this issue
by the Localized Sliced Inverse Regression (LSIR) \citep{lsir:2010}, which
takes into account the \textit{local} structure of the explanatory variables
conditioned on the response. A key observation underlying the development of
LSIR is that, especially in high dimensions, the Euclidean structure around a
data point in $\RR^p$ is only useful locally. This suggests computing a local version of the
covariance of the inverse regression $\hat{\Gamma}_{\loc}$,
which is constructed by replacing each data observation with it's smoothed
version within a local sample neighborhood with similar response values:
$$\hat \mu_{i, \loc} = \frac{1}{k} \sum_{j\in C_i} x_j,$$
where $C_i$ denotes the set of indexes of the $k$-nearest neighbors,
considering only the data points with response values within the response
slice to which $x_i$ belongs.

\subsubsection{Efficient solutions and approximate SVD}\label{sec:exact:sdr}
SIR and LSIR reduce to solving a truncated generalized eigendecomposition
problem in (\ref{eqn:gsvd:exact1}). Since we consider estimating the dimension reduction
based on sample data we focus on the sample estimators $\widehat{\Sigma}=\frac{1}{n}X^TX$
and $\widehat{\Gamma}_{XY}=X^TK_{XY}X$, where $K_{XY}$ is symmetric and encodes the
method-specific grouping of the samples based on the response $Y$.
In the classic statistical setting, when $n > p$, both
$\widehat{\Sigma}$ and $\widehat{\Gamma}_{XY}$ are positive definite almost surely.
Then, a typical solution proceeds by first sphering the data: $Z=\widehat{\Sigma}^{-\frac{1}{2}}X$,
e.g. using a Cholesky or SVD representation
$\widehat{\Sigma} = \widehat{\Sigma}^{\frac{1}{2}} (\widehat{\Sigma}^{\frac{1}{2}})^T$.
This is followed by eigendecomposition of $\widehat{\Gamma}_{ZY}$ \cite{Li:1991, lsir:2010}
and back-transformation of the top eigenvectors directions to the canonical basis.
The computational time is $O(np^2)$.
When $n<p$, $\hat{\Sigma}$ and $\hat{\Gamma}$ are rank-deficient and a
unique solution to the problem (\ref{eqn:gsvd:exact1}) does not exist. One widely-used
approach, which allows us to make progress in this problematic setting, is to restrict our attention 
to the directions in the data with positive variance. Then we can proceed as before, using 
an orthogonal projections onto the span of the data. The total computation time in this case is $O(n^2p)$.
%represent $\hat{\Sigma} \eqsvd UD^2U^T$.
%Then we apply SVD to $D^{-1}U^T\Gamma UD^{-1}$ and again back-transform to
%the canonical basis using $UD^{-1}$. The total computation time is $O(n^2p+n^3)$.
In many modern data analysis applications both $n$ and $p$ are very large, and
hence algorithmic complexity of $O[\mbox{max}(n,p) \times \mbox{min}(n,p)^2]$ could
be prohibitive, rendering the above approaches unusable.
We propose an approximate solution that explicitly recovers
the low-rank structure in $\Gamma$ using \textit{Adaptive Randomized SVD} from
Section \ref{sec:rsvd}. %Whenever there is not sufficient information to pre-sepecify 
%values for $d^{*}$ and especially $t^{*}$, they can be estimated using the data-adaptive 
%method described in Section \ref{rankest_unsup}.
In particular, assume rank$(\Gamma)=d^{*} \ge r$ (where $r$ is the dimensionality
of the optimal dimension reduction subspace).
Then $\Gamma \eqsvd US^2U^T$, where $U \in \RR^{p \times d^{*}}$.
The generalized eigendecomposition problem (\ref{eqn:gsvd:exact1}) solution
becomes restricted to the subspace spanned by the columns of $\Gamma$:
\begin{eqnarray}\label{eqn:symm:gsvd}
S^{-1}U^T \Sigma US^{-1} e &=& \frac{1}{\lambda} e, \quad \ \ \ e \equiv SU^Tg.
\end{eqnarray}
The dimension reduction subspace is contained in the $\mbox{span}(G)$, where
$G = (US^{-1} e_1,\ldots, US^{-1} e_{r})$.
%We propose to use the \textit{Adaptive Randomized SVD} algorithm from Section
%\ref{sec:rsvd} to provide a
%low-rank SVD factorization for $\Gamma$ estimating both $d^{*}$ and $t^{*}$ from
%the data.
%The intrinsic dimensionality $d$ is assumed given and its adaptive
%estimation left to future work.

%3. make explicit the application of RSVD in the context of SIR/LSIR: adaptive estimation of the
%of rank of $\Gamma$ -- r. The number of edrs is assumed to be given, if not report the 
%top r directions ordered by the estimates of the generalized eigenvalues.
%The number of 'stabl'e edrs can be estimated using  RSVD applied to 
%$\Sigma ^{-1/2}\Gamma$.

\paragraph{SIR Estimation.}
We first consider the case of SIR. Sort the samples in decreasing
order of the response and slice the samples into $H$ slices.
For each slice $h$ we define
the row vector $J_h = (1 \ \cdots \ 1)_{1\times n_h}$
and the matrix
$$J_{\text{sir}} = \small \left(
\begin{array}{ccc}
J_1 & & \mbox{\Large $0$}\\
& \ddots & \\
\mbox{\Large $0$}& & J_H
\end{array}
\right)_{n \times n}.$$
Given $J_{\text{sir}}$ and the data matrix we observe that
$$\hat{\Gamma} = X^T J_{\text{sir}}^T J_{\text{sir}} X = X^T K_{XY}^{\text{sir}} X.$$
By construction $\mbox{rank}(\hat \Gamma) \leq H-1 \ll \min(n,p)$ and
$X^T J_{\text{sir}} \in \RR^{p \times H}$ can be
constructed in $O(np)$ time. Hence, the full eigendecomposition of
$\hat{\Gamma} \eqsvd \hat{U} \hat{S}^2 \hat{U}^T$ is computable in $O(H^2p)$. Recall
that $\hat{\Sigma}=X^TX$, so the explicit construction and full eigendecomposition in
(\ref{eqn:symm:gsvd}) reduces to the exact SVD of $X\hat{U}\hat{S}^{-1}$,
resulting in $O(H^2p+Hnp)$ computation for the exact SIR solution.
When the number of slices
is small, $H \ll \mbox{min}(n,p)$,
the time savings could be substantial as compared to a typical solutions based on
the factorization of $X$ or $X^TX$ which scale as $O(n^2p)$ and O($p^3 + n^2p$),
respectively.
%In the case of randomized SIR instead of exact SVD we apply
%adaptive RSVD to estimate $\hat{U}$ and $\hat{S}$, which results in a regularized
%estimator with potentially improved statistical properties.

\paragraph{LSIR Estimation.}
We now consider the LSIR case. For each slice $h \in \{1,\ldots,H\}$
we construct a block matrix $J^{h}$ that is $n_h \times n_h$
with each entry $J^{h}_{ij} = \frac{1}{k_{h}}$ if $i,j$ are
$k$-nearest neighbors (k fixed parameter) and $0$ otherwise, $J^{h}_{ii} = \frac{1}{k_{h}}$,
$k_{h} \ge k$. We use a symmetrized version of the $k$-nearest neighbors (kNN), which postulates
that two entries are neighbors of each other if either one is within the
$k$-nearest neighborhood of the other. Hence,
$k_{h} = \mbox{number of neighbors after symmetrization}$.
This results in symmetric $\hat \Gamma_{\loc}$.
We then construct the block diagonal $n \times n$ matrix
$J_{\text{lsir}} = \mbox{diag}(J_1,...,J_H)$. Then
$$\hat \Gamma_{\loc} = X^T J_{\text{lsir}}^TJ_{\text{lsir}}X = X^T K_{XY}^{\text{lsir}} X.$$
To compute the kNN we need the pairwise distances between points.
When $n \ll p$ we could use an approximate kNN, which applies
exact kNN on the project the data using Gaussian Random projection.
Based on the Johnson-Lindenstrauss Lemma \citep{Johnson1984} and more recent
related work \citep{JLS:Frankl:1987, Achlioptas:2001},
we expect to incur a minimal loss in accuracy at the random projection step as
long as the dimensionality of the projection space is $O(log(n))$. Hence we fix that
dimension to be $k=20 \times log_2(n)$ and use $X_{\Omega} = X \Omega^{*} \in \RR^{n \times k}$
(instead of $X$) in the kNN search. Here
$\Omega^{*} \in \RR^{p \times k}$ is the random projection matrix, which
is generated in two steps:
\begin{enumerate}
\item generate random matrix: $\Omega_{ij} \iid N(0, 1)$, $\Omega \in \RR^{p \times k}$
\item calculate orthonormal basis: $U_{\Omega} D_{\Omega} V^T_{\Omega} \eqsvd \Omega$, set $\Omega^{*} \equiv U_{\Omega}$.
\end{enumerate}
Constructing $X^T J_{lsir} \in \RR^{p \times n}$ requires
$O(d^{*}np)$ operations. The full eigendecomposition of
$\hat{\Gamma}_{\loc}$, requires $O(n^2p)$ operations, which is followed
by the final step of finding the top eigenvalues and eigenvectors of
(\ref{eqn:symm:gsvd}) using exact SVD and back-transforming into
the ambient basis. That step takes $O(d^{*}np)$. Hence the runtime
complexity of LSIR becomes $O(n^2p + d^{*}np)$. For the randomized
estimator of LSIR we use \textit{Adaptive RSVD} to approximate $\hat{\Gamma}_{\loc}$,
which results in reduced overall runtime complexity of $O(td^{*}np)$.
A detailed description of the algorithms is contained in the
Appendix section, Algorithm \ref{alg:RSIR_LSIR}.

\section{Results on real and simulated data}\label{sec:results}
We use real and simulated data to demonstrate three major points: 
\begin{enumerate}
\item In the presence of interesting \textit{low-rank} structure in the data, 
the randomized algorithms tend to be much faster than the exact methods 
with minimal loss in approximation accuracy.

\item The \textit{rank} and the \textit{subspace} containing the information 
in the data can be reliably estimated and used to provide efficient
solutions to dimension reduction methods based on the truncated (generalized) 
eigendecompositon formulation.

\item The randomized algorithms allow for
added regularization which can be adaptively controlled in computationally
efficient manner to produce improved out-of-sample performance.
\end{enumerate}

\subsection{Simulations}
\subsubsection{Unsupervised dimension reduction}\label{sim:unsup}
We begin with unsupervised dimension reduction of data with low-rank
structure contaminated with Gaussian noise and focus on evaluating the
application of \textit{Adaptive Randomized SVD} (see Section \ref{sec:rsvd})
for PCA. In particular, we demonstrate that the proposed method estimates the
sample singular values with exponentially decreasing relative error in $t$.
Then we show that achieving similar low-rank approximation accuracy to a
state-of-the-art Lanczos method requires the same runtime complexity,
which scales linearly in both dimensions of the input matrix. This makes our 
proposed method applicable to large data. Lastly, we demonstrate the ability 
to adaptively estimate the underlying rank of the data, given a coarse upper 
bound. For the evaluation of the randomized
methods, based on all our simulated and real data, we assume a default
value for the oversampling parameter $\Delta=10$.

\paragraph{Simulation setup}
The data matrix $X \in \RR^{n \times p}$, is generated as follows:
$X=USV^T+E$, where $U^TU=V^TV=I_{d^{*}}$. The $d^{*}$ columns
of $U$ and $V$ are drawn uniformly at random from the corresponding
unit sphere and the singular values S = $\mbox{diag}(s_1,\ldots,s_{d^*})$
are randomly generated starting from a baseline value,
which is a fraction of the maximum noise singular value,
with Exponential increments separating consecutive entries:
\begin{eqnarray*}\label{eqn1_sim:unsup}
s_j &=& \nu_{j-1} + \nu_{j}, \ \mbox{for}\ j \in \{1,\ldots,d^{*}\} \\
\nu_j &\iid& \mbox{Exp}(1), \ \ \ \nu_0 = \kappa \times s^{(E)}_1.
\end{eqnarray*}
The noise is iid Gaussian: $E_{ij} \iid N(0, \frac{1}{n})$. The sample
variance has the SVD decomposition $E \eqsvd U_E S_EV^T_E$, where
$S_E = \mbox{diag}(s^{(E)}_1,\ldots,s^{(E)}_{min(n,p)})$ are the
singular values in decreasing order.
The signal-to-noise relationship is controlled by $\kappa$,
large values of which correspond to increased separation
between the signal and the noise.

\paragraph{Results.}
In our first simulation we set $\kappa=1$ and assume the
rank $d^{*}$ to be given and fixed to $50$. The main focus
is on the effect of the regularization parameter $t$ controlling
the singular value shrinkage, with larger
values corresponding to stronger preference for the higher
vs. the lower variance directions. The simulation uses an
input matrix of dimension $2,000 \times 5,000$.
In Table \ref{tab1} we report the estimates o the 
\% relative error of the \textit{singular values} averaged over 
10 independent random data sets. We clearly observe exponential convergence 
to the sample estimates with increasing $t$. This suggests that the variability
in the sample data directions can be approximated well for very
large data sets at the cost of few data matrix multiplications.
%y = 100-c(74.3, 91.2, 97.1, 99.1, 99.7)
%y.se = c(0.5,0.6,0.4,0.2,0.1)
%x = 1:length(y)
%require(gplots)
%plot(x,y, log="y", xlab = "t", ylab = "relative error (%)", cex=0.6)
%plotCI(x, y, uiw=2*y.se, log="y", lwd=0.6, pch=NA, cex=0.6, slty=1, sfrac=0.005, type="p", gap=0, add=T)
%abline(lm(log10(y)~x), col="red", lty=2)
%%%%%%%%
%% Table 1
%%%%%%%%
\begin{table}[ht]
\begin{center}
\begin{tabular}{|c|ccccc|}
\hline
t & 1 & 2 & 3 & 4 & 5 \\
\hline
%%\% relative accuracy & $72.8 \pm 0.2$ & $86.9 \pm0.2$ & $94.1 \pm 0.2$ & $97.5 \pm 0.2$ & $ 99.0 \pm 0.1$\\
%%$log_2$(\% relative accuracy) & $3.2$ & $2.7$ & $1.1$ & $-0.1$ & $ -1.2 $\\
%% log(100-accuracy) = 3.2425924 2.1747517 1.0647107 -0.1053605 -1.2039728 [LINEAR trend]
\% relative error & $26.1\pm 0.2$ & $8.8\pm 0.2$ & $3.0\pm 0.1$ & $1.0\pm 0.1$ & $0.3\pm 0.0$\\
$\mbox{log}_{10}$[\mbox{mean}] & $4.7$ & $3.1$ & $1.6$ & $-0.1$ & $-1.9$ \\
\hline
\end{tabular}
\caption{ Singular value estimation using Randomized SVD. We report 
the mean relative error ($\pm 1$ standard error) averaged across 10 random 
replicates of dimension $2,000 \times 5,000$, rank 50. Linear increase in 
the regularization parameter $t$ results in exponential decrease in the 
\% relative error.}
\label{tab1}
\end{center}
\end{table}

\noindent Similar ideas have been proposed before in the absence of
randomization. Perhaps, the most well-established and computationally
efficient such methods are based on Lanczos Krylov Subspace
estimation, which also operate on the data matrix only through
matrix multiplies \citep{saad1992numerical, arpack, stewart2001matrix, irlba}.
All these methods are also iterative in nature and tend to converge
quickly with runtime complexity, typically, scaling as the product of the
data dimensions $O(qnp)$, with $q$ -- small.

In order to further investigate the practical runtime behavior of
\textit{Adaptive Randomized SVD} we used one such state-of-the art
low-rank approximation algorithm, Blocked Lanczos \citep{irlba}. Here
we use the same simulation setting as before with fixed $d^{*}=50$, but vary the
regularization parameter $t$ to achieve comparable \% relative low-rank
Frobenius norm reconstruction error (to 1 d.p). In
Table \ref{tab2} we report the ratio of the runtimes of the two
approaches based on 10 random data sets.
Notice that the relative runtime remains approximately constant with simultaneous
increase in both data dimensions, which suggests similar order of complexity for both
methods. This results suggest that, in the presence of low-rank structure,
\textit{Adaptive Randomized SVD} would be feasible for larger data problems
than full-factorization methods e.g. \citep{Anderson:1990:LAPACK},
achieving good approximation accuracy for the top sample singular
values and singular vectors.
%%%%%%%%
%% Table 2
%%%%%%%%
\begin{table}[ht]
\begin{center}
\begin{tabular}{|r|ccccc|}
\hline
n + p & 6,000 & 7,500 & 9,000 & 10,500 & 12,000\\
\hline
relative time & $2.5 \pm 0.05$& $1.84 \pm 0.03$& $1.82 \pm 0.03$& $1.83 \pm 0.02$ & $1.84 \pm 0.04$\\
\hline
\end{tabular}
\caption{Runtime ratio between \textit{Adaptive Randomized SVD} and Lanczos for
different data dimensions $n$ and $p$ and rank $d^{*}=50$. The Lanczos implementation is
from the 'irlba' CRAN package \citep{irlba}. The \textit{Adaptive Randomized SVD}
was run until the \% relative low-rank reconstruction error was equal to the Lanczos error
to 1 d.p. For each consecutive simulation scenario $n$ is incremented by 500 and $p$ is
 incremented by 1,000. We report the sample mean and standard error based on 10 random 
replicate data sets.}
\label{tab2}
\end{center}
\end{table}

\begin{figure}[h!tb]
\begin{minipage}{.4\textwidth}
\centering
\includegraphics[scale=0.42]%%{./output/unsupervised/rank_estimation/num_rand=100_exp_mean=1/nfolds=2_num_cv=1_num_q=5/fixed_rank_upper_bound=80/kappa=2/n=2000_p=5000_k_min=10_k_max=50_d_hat.pdf}
{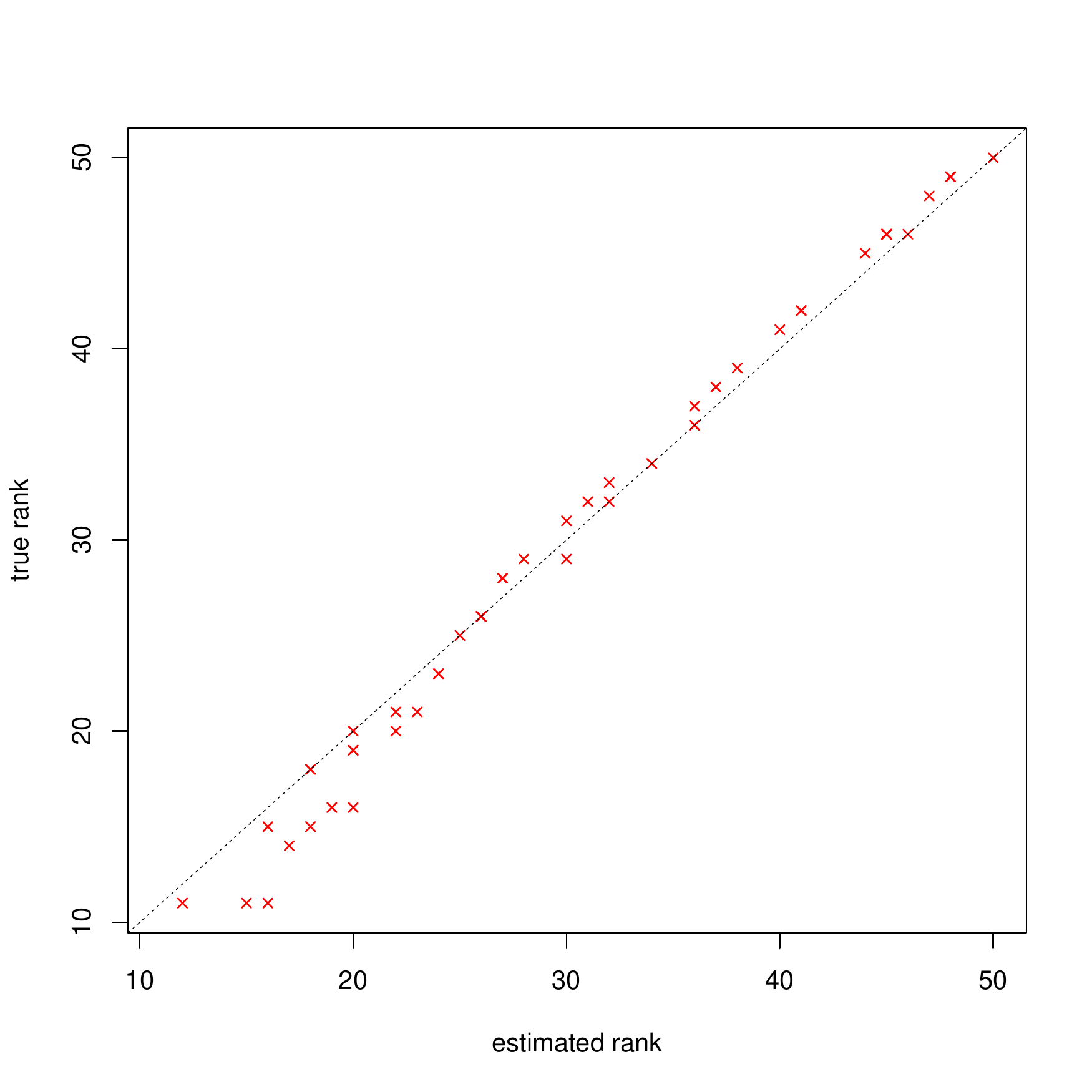}
\end{minipage}\qquad \qquad \qquad
\begin{minipage}{.4\textwidth}
\centering
\includegraphics[scale=0.42]%%{./output/unsupervised/rank_estimation/num_rand=100_exp_mean=1/nfolds=2_num_cv=1_num_q=5/fixed_rank_upper_bound=80/kappa=2/n=2000_p=5000_k_min=10_k_max=50_t_hat.pdf}
{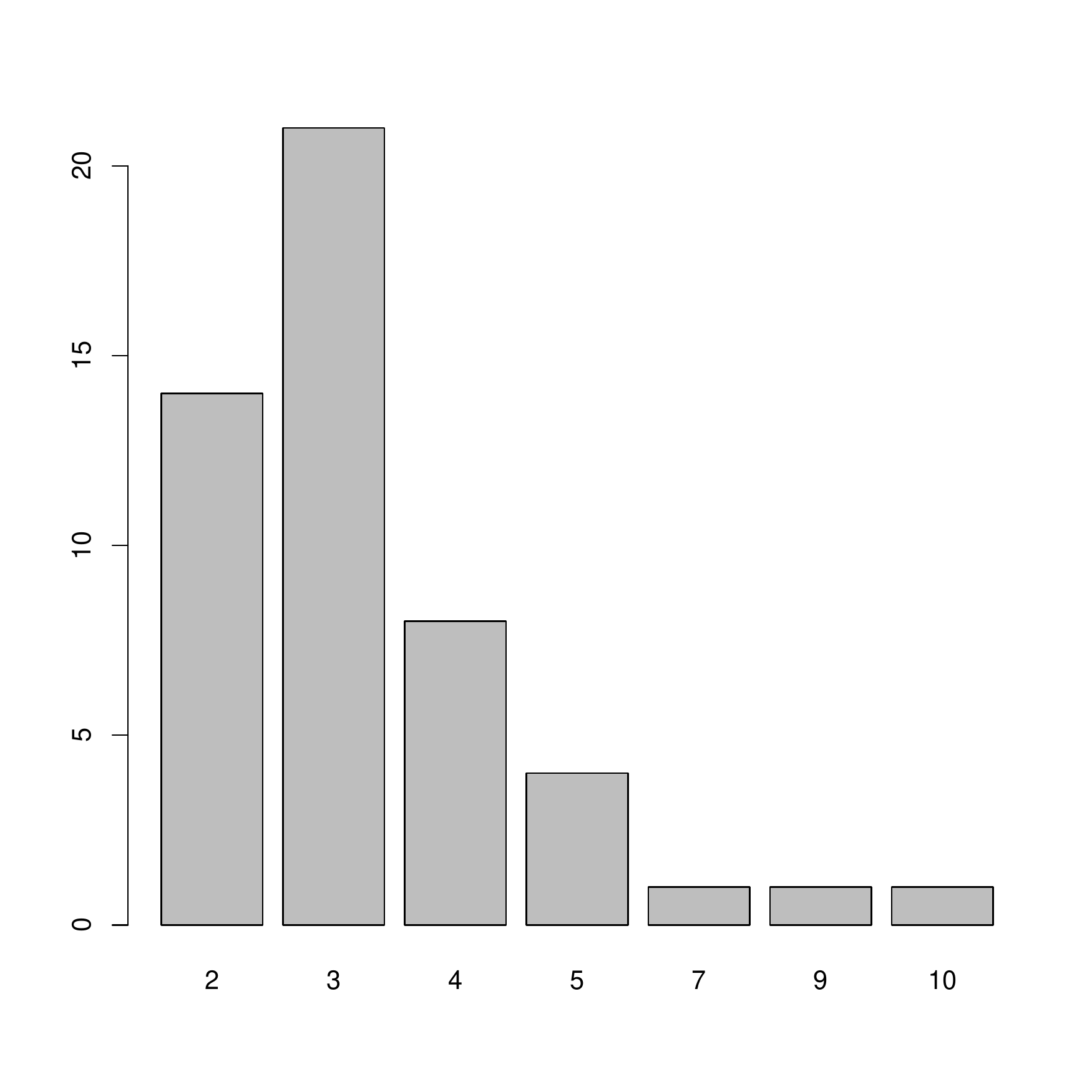}
\end{minipage}
\caption{Simulation results from 50 random data sets of dimension n=2000, p=5000
, $d^{*} \iid \mbox{Uniform}[10,50]$, $\kappa=2$.
Left panel reports the estimates of $\hat{d}^{*}$ vs the true rank $d^{*}$. The right panel reports the selected vaues for $t^{*}$.}
\label{fig:rank_t_kappa2}
\end{figure}

\begin{figure}[h!tb]
\begin{minipage}{.4\textwidth}
\centering
\includegraphics[scale=0.42]%%{./output/unsupervised/rank_estimation/num_rand=100_exp_mean=1/nfolds=2_num_cv=1_num_q=5/fixed_rank_upper_bound=80/kappa=1/n=2000_p=5000_k_min=10_k_max=50_d_hat.pdf}
{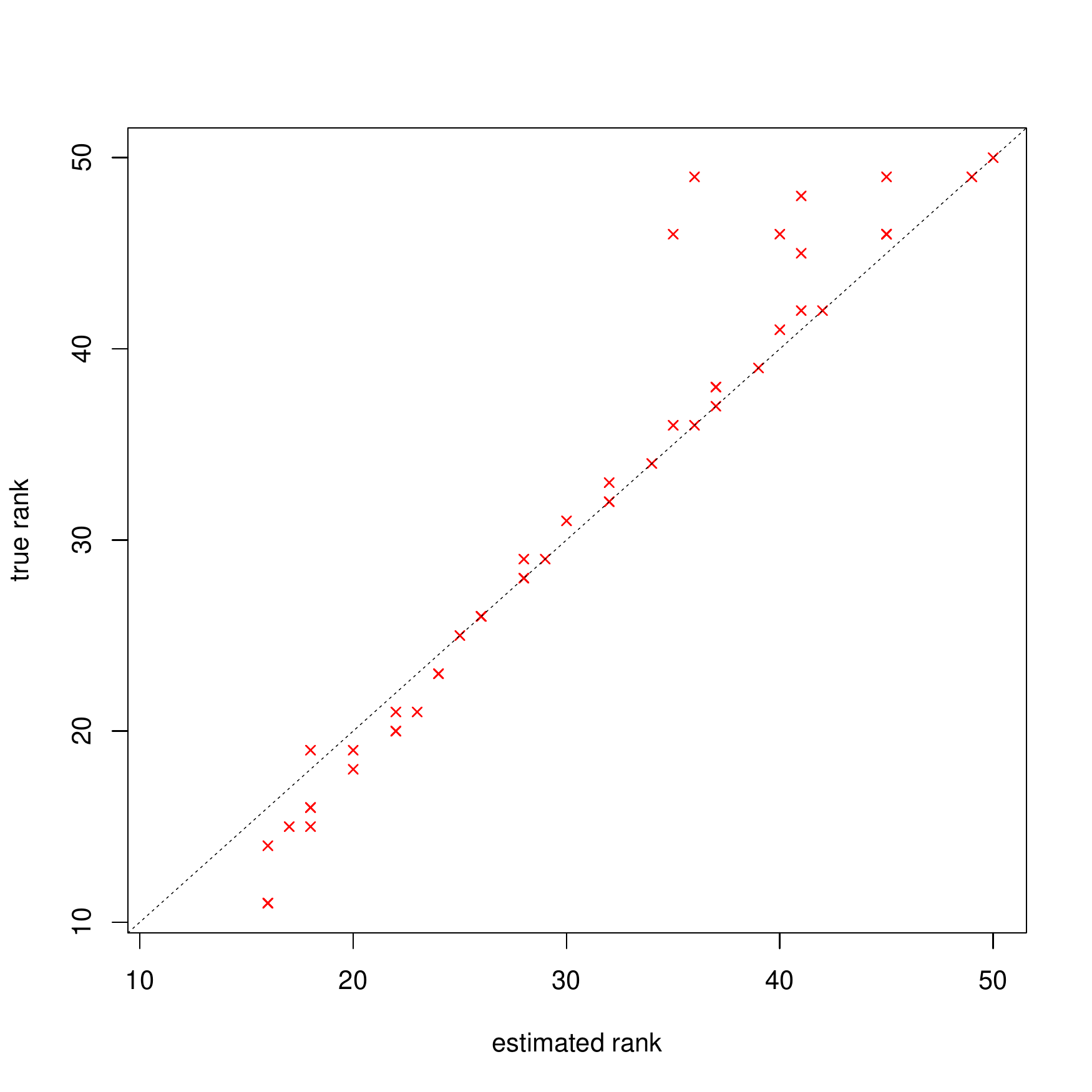}
\end{minipage}\qquad \qquad \qquad
\begin{minipage}{.4\textwidth}
\centering
\includegraphics[scale=0.42]%%{./output/unsupervised/rank_estimation/num_rand=100_exp_mean=1/nfolds=2_num_cv=1_num_q=5/fixed_rank_upper_bound=80/kappa=1/n=2000_p=5000_k_min=10_k_max=50_t_hat.pdf}
{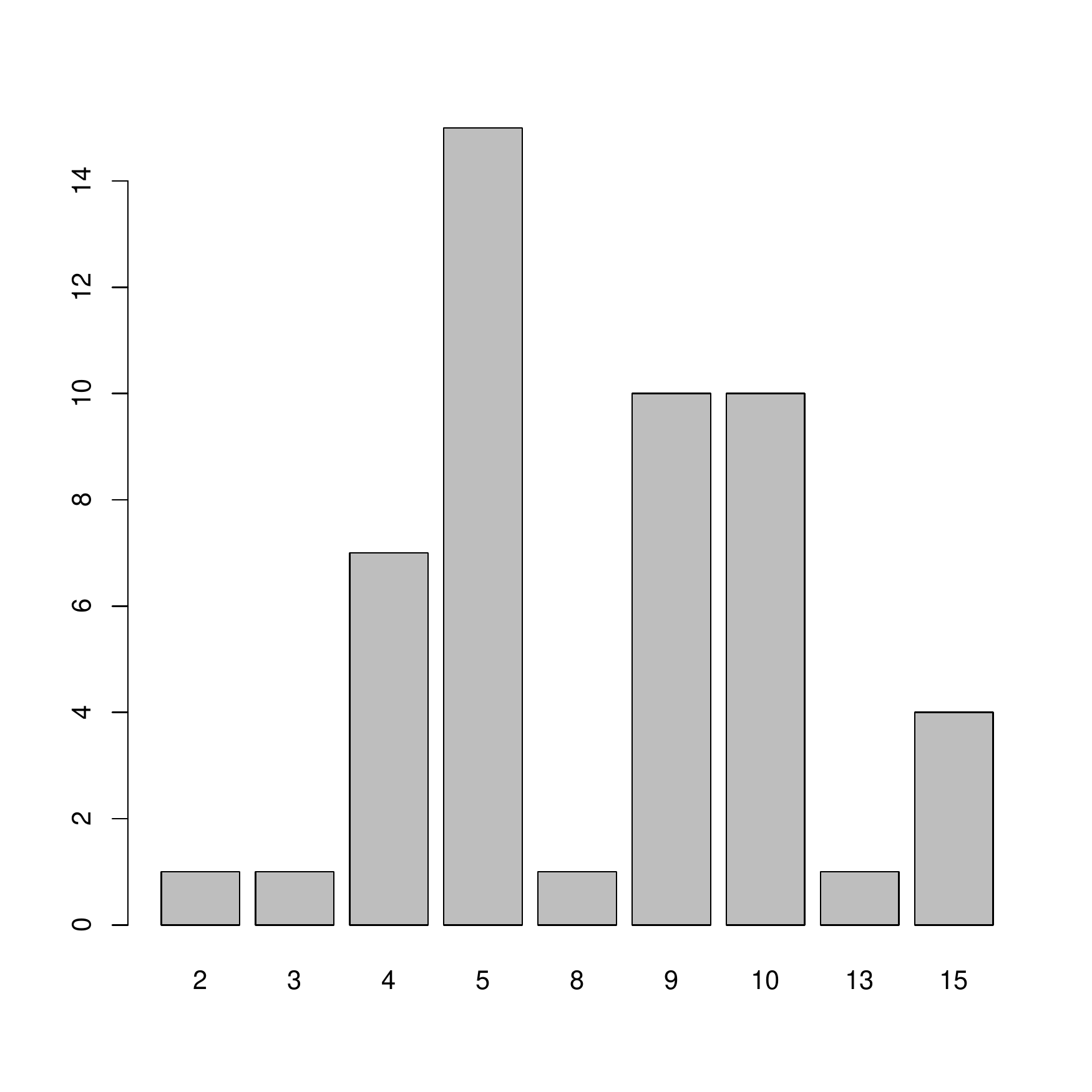}
\end{minipage}
\caption{Simulation results from 50 random data sets of dimension n=2000, p=5000
, $d^{*} \iid \mbox{Uniform}[10,50]$, $\kappa=1$. Left panel reports the estimates of
$\hat{d}^{*}$ vs the true rank $d^{*}$. The right panel reports the selected vaues for $t^{*}$.}
\label{fig:rank_t_kappa1}
\end{figure}

In many modern applications the data have latent underlying
low-rank structure of unknown dimension. In Section \ref{rankest_unsup} we address the
issue of estimating the rank using a stability-based approach. Next we study the
ability of our randomized method for rank estimation to identify the
dimensionality in simulated data of the low-rank. For that purpose we generate 50 random
data sets, $n=2,000, p= 5,000$, $d^{*} \iid \mbox{Uniform}[10,50]$,
and $\kappa=1, 2$. We set an initial rank upper bound estimate to be
$d^{*} + \mbox{delta}$ (where $\mbox{delta}=30$) and use the adaptive
method from Section \ref{rankest_unsup} to estimate both optimal $t^{*}$ and 
the corresponding $d^{*}$. Figure \ref{fig:rank_t_kappa2} and \ref{fig:rank_t_kappa1} 
plots the \textit{estimated} vs. the \textit{true rank} and the corresponding estimates of the
regularization parameter $t^{*}$ for two ``signal-to-noise" scenarios. In both cases the 
rank estimates show good agreement with the true rank values. When $\kappa=1$
the smallest "signal" direction has the same variance as the largest variance "noise" 
direction, which causes our approach to slightly underestimate the largest ranks. 
This is due to the fact that the few smallest variance signal directions tend to be 
difficult to distinguish from the random noise and hence less stable under random 
projections. We observe that our approach tends to select small values for $t^{*}$, 
especially when there is a clear separation between the signal and the noise
 (right panel of Figure \ref{fig:rank_t_kappa2}).
This results in reduced number of matrix multiplies and hence fast computation.
%The sensitivity to the value of $\mbox{delta}$ was investigated by plotting the
%estimated vs. observed rank for different values of $\mbox{delta}$. Figure
%\ref{fig:rank_kappa_delta} contains results from 50 simulated data sets under the 
%same parameters used for the rank estimation, but varying the value for$\mbox{delta}$. 
%For both values of $\kappa$, we observe some sensitivity to the choice of $delta$. In general
%larger values tend to result in overestimation of the rank. In this specific simulation setup
%we observe particularly good performance for $delta=30$. Further investigation of this
%problem is left to future work. 
%%%%%%%%
% Figure 1
%%%%%%%%
%%%%%%%%
% Figure 2
%%%%%%%%
%%%%%%%%
% Figure 3
%%%%%%%%
%\begin{figure}[h!tb]
%\begin{minipage}{.4\textwidth}
%\centering
%\includegraphics[scale=0.47]%%{./output/unsupervised/rank_estimation/num_rand=100_exp_mean=1/nfolds=2_num_cv=1_num_q=5/fixed_rank_upper_bound=80/kappa=1/n=2000_p=5000_k_min=10_k_max=50_d_hat.pdf}
%{./output/unsupervised/rank_estimation=stab/num_rand=50_exp_mean=1/num_q=8_ties_method=min/kappa=1/n=2000_p=5000_k_min=10_k_max=50_rub_buffer_curves.pdf}
%\end{minipage}\qquad \qquad \qquad
%\begin{minipage}{.4\textwidth}
%\centering
%\includegraphics[scale=0.47]%%{./output/unsupervised/rank_estimation/num_rand=100_exp_mean=1/nfolds=2_num_cv=1_num_q=5/fixed_rank_upper_bound=80/kappa=1/n=2000_p=5000_k_min=10_k_max=50_t_hat.pdf}
%{./output/unsupervised/rank_estimation=stab/num_rand=50_exp_mean=1/num_q=8_ties_method=min/kappa=2/n=2000_p=5000_k_min=10_k_max=50_rub_buffer_curves.pdf}
%\end{minipage}
%\caption{Estimation of $d^*$, given upper bound estimate $d_{max} = d^{*} + \mbox{delta}$. We report simulation
%results from 50 random data sets of dimension n=2000, p=5000, $d^{*} \iid \mbox{Uniform}[10,50]$.
%The signal-to-noise is set to be $\kappa=1$ (left panel) and $\kappa=2$ (right panel). The plotted lines
%show loess fit through the points with coordinates $\widehat{d}^{*}$ and $d^{*}$ for different values of delta.}
%\label{fig:rank_kappa_delta}
%\end{figure}

\subsubsection{Supervised dimension reduction}\label{sim:sup}
\paragraph{Classification example.}\label{simulation:model:xor}
This section demonstrates the performance of the dimension reduction
approaches on the XOR classification example. Our particular simulation setup 
is a generalization from \citep{lsir:2010}, Section 4. The first $d^*$ dimensions of the simulated
data contain axis-aligned signal which is centered around $(\pm 2,\pm 2)$ and 
symmetric about 0 with added isotropic Gaussian noise. All remaining
dimensions contain only Gaussian noise (see Figure \ref{fig:xor_eq} in the
Appendix for an example of $d^*=4$).
In order to study the classification performance of the dimension reduction
methods we generate data with $d^*=10$ and for $i \in \{1,\ldots,n\}$:
\begin{eqnarray}
X_{ij}&=&\nu_i + \pi \delta_{-2}(X_{ij}) + (1-\pi)\delta_{2}(X_{ij}), \ \ \ \mbox{for} \ j \in \{1,\ldots,d^{*}\}\\
X_{ij}&=&\nu_i, \ \ \ \mbox{for} \ j > d^*\\
\nu_i &\iid& N(0, \sigma),
\end{eqnarray}
where $\pi=0.5$ and $\sigma$ controls the signal-to-noise relationships.
We investigate the classification performance of $k$-nearest neighbor
classifier applied to the projected data onto the estimates of the
dimension reduction directions based on different reduction methods.
To evaluate the classification accuracy we average across 
values for $k$ ranging from
10\% to 30\% of the number of samples within each of the 10 distinct clusters
(step 2 between consecutive values).
In the case of LSIR we set the number of nearest neighbors for the smoothing
step to be 20\% of the samples within each cluster.
The methods we focus on are: SIR and LSIR as implement by the (randomized)
Algorithm \ref{alg:RSIR_LSIR}, denoted as \textit{sir} and \textit{lsir}, respectively.
Whenever randomization is used we refer to the LSIR methods as \textit{rand.lsir}.
%The classic SIR as implemented in the dr CRAN package \citep{Weisbergsoft} and
The Gaussian Random projections, \textit{rand.proj}, and \textit{oracle} (dimension
reduction directions known) are included to serve as a reference.

\begin{figure}[h!tb]
\begin{minipage}{.4\textwidth}
\centering
\includegraphics[scale=0.47]
{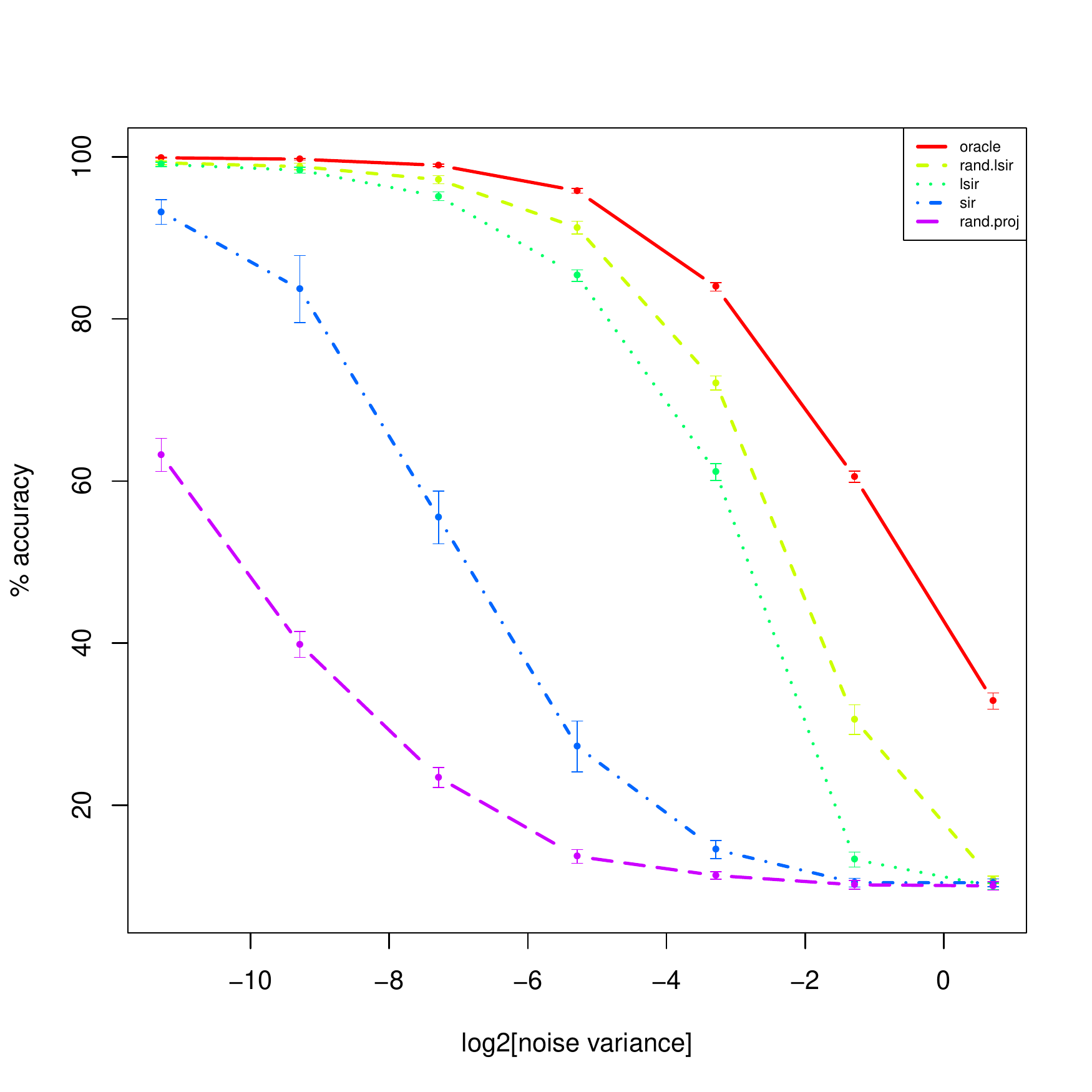}
\end{minipage}\qquad \qquad \qquad
\begin{minipage}{.4\textwidth}
\centering
\includegraphics[scale=0.47]
{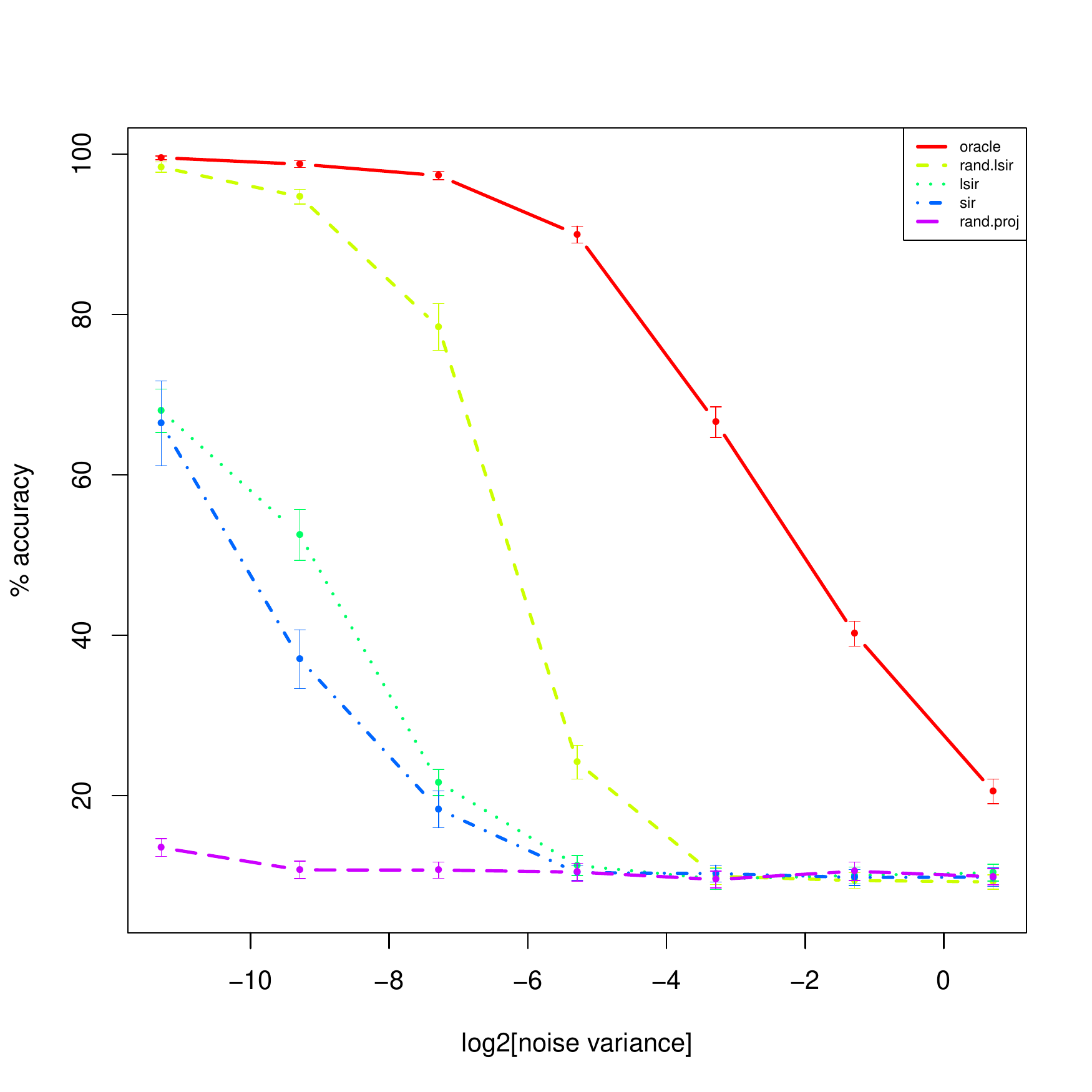}
\end{minipage}
\caption{Simulation results from 20 random data sets with signal dimension of XOR $d^{*}=10$.
The left panel reports the estimates for ambient dimension is $n=1000, p=200$. The right panel
reports the estimates for $n=200, p=1000$. Error bars are based on 2 standard errors of the
sample mean.}
\label{fig:xor_sigma}
\end{figure}

\paragraph{Results.}
Figure \ref{fig:xor_sigma} illustrates the results
when $p \ll n$ and $n \ll p$, respectively, focusing on the estimation
accuracy for different values of the variance of the Gaussian noise $\sigma^2$
(on logarithmic scale).
%As expected, dr.sir performs the worst, but perhaps surprisingly sir does a
%reasonably good job at preserving the proximity of the class members for
%smaller levels of the noise.
LSIR without randomization performs better than the
SIR methods, as expected, which is particularly evident in the data-rich scenario.
LSIR with randomization performs better than all other methods in both sample size 
regimes, with a particularly large improvement in the data-poor scenario. In
\citep{lsir:2010}, Section 4, LSIR with ridge regularization performed well in a
similar setting, however here we observe that \textit{rand.lsir} performs well even
without adding such term.

%%%%%%%%
% Figure 4
%%%%%%%%

\paragraph{Latent factor regression model}\label{sim:lfr}
In this section we explore the accuracy of randomized dimension reduction methods in
the setting of factor regression. The modeling assumption is that the response
variable $Y$ is strongly correlated with a few directions in covariate space $X$. We
compare the proposed estimators with and without randomization. In the latter
case we apply full SVD factorization to the input matrices.
The methods we focus on are: SIR and LSIR as implement by the (randomized)
Algorithm \ref{alg:RSIR_LSIR}, denoted as \textit{sir} and \textit{lsir}, respectively.
Whenever randomization is used we refer to the methods as \textit{rand.sir}
and \textit{rand.lsir}. PCA is denoted as \textit{pca}, and (randomized) PCA as
implemented by the Algorithm \ref{alg:arsvd} by \textit{rand.pca}.
We will use a latent
factor regression model \citep{West03} to simulate the data used to evaluate
the performance of the different methods. The rationale behind using a latent
factor regression model is that we have explicit control of the direction of variation
in covariates space $X$ which is strongly correlated with the response $Y$.
The latent factor model corresponds to the following decomposition
\begin{eqnarray*}
Y_i &=& \lambda_i^T \theta + \epsilon_i,\\
X_{i} &=& BS\lambda_i + \nu_i,
\end{eqnarray*}
where $B \in \RR^{p \times d^*}$, $B^TB=I_{d^*}$, and $S=\mbox{diag}(s_1,\ldots,s_{d^*})$ and $d^* \ll p$.
The random variables $\lambda_i, \nu_i, \epsilon_i$ are independent and normally distributed
\begin{eqnarray*}
\vecthree{\lambda_i}{\epsilon_i}{\nu_i} \sim \mbox{N}(0,V), \ \ V = \diagthree{I_{d^*}}{\tau^2}{\psi^2 I_p}.
\end{eqnarray*}
Under the above model as the noise in the covariates decreases ($\psi^2 \rightarrow 0$), we obtain
\begin{eqnarray*}
Y_{i} \mid x_{i} & \sim& \mbox{N}(x_i^TBS^{-1} \theta, \tau^2),
\end{eqnarray*}
which corresponds to the principal components regression model
$$Y_i \mid x_i \sim \mbox{N}(\tilde{x}_i \tilde{\theta},\tau^2), \quad \mbox{with }\tilde{x}_i= B^T x_i  \mbox{ and } 
\tilde{\theta} = S^{-1}\theta.$$
The dependence between $X_i$ and $Y_i$ is induced by marginalizing the latent factors
$\lambda_i$. The joint distribution for $(Y_i,X_i)$ is normal
$$	\vectwo{X_i}{Y_i} \sim \mbox{N}(0,\Sigma), \ \ \Sigma = \mattwo{BS^2B^T + \psi^2 I}{BS\theta}{\theta^TSB^T}{\tau^2 + \theta^T\theta},$$
as is the conditional
$$Y_i \mid x_i, \theta, S, \sim \mbox{N}\left(\theta^T S B^T C x_i, \tau^2 + \theta^T(I - SB^TCBS)\theta\right), \quad C=(BS^2B^T + \psi^2 I)^{-1} \in \RR^{p}.$$
For the above model the parameters $\psi^2$ and $\tau^2$ control the percentage
of variance explained or signal-to-noise
$$\mbox{s}2\mbox{n}_x = \frac{\min(s_i^2)}{\psi^2+\min(s_i^2)} , \quad \mbox{s}2\mbox{n}_y= \frac{\theta^T\theta}{\tau^2+\theta^T\theta}=\frac{\mbox{Var}(\lambda^T_i\theta)}{\mbox{Var}(Y_i)}.$$
We focus on two signal-to-noise regimes, setting both parameters
to vary uniformly at random between (A) 0.6 and 0.9 -- strong signal, and
(B) 0.3 and 0.6 -- weak signal.
In both simulation scenarios we set the dominant directions of variation in
the covariates to correlate with the regression coefficient. In particular, the parameters
$\{\theta_i\}$ and $\{s_i\}$ are drawn from a t-distribution with 5 degrees
of freedom, with the eigenvalue directions assigned such that
$|\theta_1| > |\theta_2| > \ldots > |\theta_{d^*}|$ and $|s_1| > |s_2| > \ldots > |s_{d^*}|$.
The the number of factors is random, $d^* \iid \mbox{Uniform}[5,20]$, and the
covariance structure is set to be spherical $\psi^2 I$.

\paragraph{Evaluation criteria.}
We used three criteria to evaluate the estimates of the projection
direction we obtained from the different methods.
The first was the absolute value of the correlation of the dimension
reduction space which in our simulations was the vector
$$b = S_{Y|X}=\mbox{span}[\cov(X)^{-1}\cov(X,Y)]=\mbox{span}[(BS^2B^T
+ \psi^2 I)^{-1}BS\theta].$$
We report the absolute correlation (AEDR) of $b$ with the effective dimension
reduction estimate $\hat{b}$, $|\mbox{corr}(b,\hat{b})|$. The second
and third metric are based on predictive criteria. In the first case the
criterion used is an estimate of the mean square prediction error (MSPE)
$$E \|Y-Z \hat{b} \|^2_2,$$
where $Z$ is the projection of the data $X$ onto the edr subspace and
$\hat{b}$ are the regression coefficients estimates. The
third metric is the proportion of the variance explained
by the linear regression ($R^2$), $\mbox{Cor}(Y,\hat{Y})^2$, where
$\hat{Y} = Z \hat{b}.$
We generated $20$ data sets. The performance on the different
evaluation criteria was estimated using test data generated from the
same simulation setup as the training data:
\begin{eqnarray*}
\lambda^{*}_i \iid N(0, I_{d^*}), \  \epsilon^{*}_{i} \iid N(0, \tau^2),\ \nu^{*}_{i} \iid N(0, \psi^2 I_p)\\
Y_{test,i} = \lambda_{i}^{*T} \theta + \epsilon^{*}_{i}, \ \ \ X_{test,i} = B S \lambda^{*}_{i} + \nu^{*}_{i}.
\end{eqnarray*}
For each of the two parameter settings we examined two data size
regimes, $n > p$ and $p > n$. For the LSIR and SIR algorithms
the number of slices is set to ten, $H=10$, and for LSIR the
nearest neighbor parameter is set to ten, $k=10$. The rank of
edr subspace is set for all algorithms to the true value $r=d^*=1$. For all
randomized methods we use the adaptive strategy described in
Section \ref{sec:adapt_t} to estimate the value for $t^*$.

\paragraph{Results.}
Table \ref{fr:nlarge} and \ref{fr:plarge} report the results
for the various dimension reduction approaches methods in the case of high signal-to-noise
($\mbox{s}2\mbox{n}_x, \mbox{s}2\mbox{n}_y \iid \mbox{Uniform}[0.6,0.9]$)
and low signal-to-noise 
($\mbox{s}2\mbox{n}_x, \mbox{s}2\mbox{n}_y \iid \mbox{Uniform}[0.3,0.6]$)
for both $n >p$ and $n < p$. In all cases the supervised 
methods outperform PCA-based methods. Interestingly, randomization tends 
to be beneficial in the data-poor regime, irrespective of the signal-to-noise ratio.
In the setting where the regression
signal is low compared to the noise and the sample size is small
(Table \ref{fr:plarge}), randomization seems to have
the strongest positive effect. It improves both the
predictive performance as well as the estimate of the true edr for SIR and LSIR.
It is known that LSIR tends to perform much better with added regularization
term \citep{lsir:2010}. But we observe that factorizing $\hat{\Gamma}$
instead of $\hat{\Sigma}$, which is the typical approach adopted in other
LSIR solutions, seems to alleviate this problem. This seems to be the case in the
data-rich regime, where lsir performs well without extra regularization. %Perhaps, the randomization adds an
%implicit regularization in this case.
%%%%%%%%
%% Table 3
%%%%%%%%
\begin{table}[ht]
\centering
\begin{tabular}{|l|ccc|ccc|}
\hline
%&\multicolumn{3}{|c|}{$\mbox{s}2\mbox{n}_x = \mbox{s}2\mbox{n}_y \in [0.3,0.6]$}
%& \multicolumn{3}{|c|}{$\mbox{s}2\mbox{n}_x = \mbox{s}2\mbox{n}_y \in [0.6,0.9]$} \\
&\multicolumn{3}{|c|}{Low signal-to-noise}
& \multicolumn{3}{|c|}{High signal-to-noise} \\
\cline{2-7}
Method & $R^2$ & MSPE & AEDR & $R^2$ & MSPE & AEDR \\
\hline
sir & $\bf{0.45\pm0.02}$ & $\bf{1.04\pm0.01}$ & $0.54\pm0.05$ & $\bf{0.75\pm0.02}$ & $\bf{1.16\pm0.03}$ & $\bf{0.56\pm0.05}$ \\
rand.sir & $0.37\pm0.02$ & $1.20\pm0.03$ & $\bf{0.57\pm0.04}$ & $0.57\pm0.03$ & $2.12\pm0.16$ & $0.53\pm0.05$ \\
lsir & $0.33\pm0.02$ & $1.27\pm0.03$ & $0.10\pm0.02$& $0.64\pm0.03$ & $1.67\pm0.12$ & $0.18\pm0.03$ \\
rand.lsir & $0.22\pm0.02$ & $1.49\pm0.05$ & $0.35\pm0.04$ & $0.40\pm 0.03$ & $2.97\pm 0.22$ & $0.35\pm 0.04$ \\
pca & $0.19\pm0.02$ & $1.56\pm0.06$ & $0.29\pm0.04$ & $0.32\pm0.03$ & $3.48\pm0.29$ & $0.27\pm0.03$ \\
rand.pca & $0.19\pm0.01$ & $1.56\pm0.05$ & $0.30\pm0.04$ & $0.28\pm0.03$ & $3.69\pm0.31$ & $0.24\pm0.03$ \\
\hline
\end{tabular}
\caption{Estimates of the mean performance and its standard error based on the Latent Factor Regression model. 
The results are based on 20 replicate data sets of dimension $n=3000$, $p=500$.}
\label{fr:nlarge}
\end{table}
%%%%%%%%
%% Table 4
%%%%%%%%
\begin{table}[ht]
\centering
\begin{tabular}{|l|ccc|ccc|}
\hline
%&\multicolumn{3}{|c|}{$\mbox{s}2\mbox{n}_x = \mbox{s}2\mbox{n}_y \in [0.3,0.6]$}
%& \multicolumn{3}{|c|}{$\mbox{s}2\mbox{n}_x = \mbox{s}2\mbox{n}_y \in [0.6,0.9]$} \\
&\multicolumn{3}{|c|}{Low signal-to-noise}
& \multicolumn{3}{|c|}{High signal-to-noise} \\
\cline{2-7}
Method & $R^2$ & MSPE & AEDR & $R^2$ & MSPE & AEDR \\
\hline
sir & $0.15 \pm 0.05$ & $1.58 \pm 0.10$ & $ 0.16\pm 0.05$ & $0.32 \pm0.08$ & $3.23\pm0.49$ & $0.26\pm0.07$\\
rand.sir & $\bf{0.34 \pm0.02}$& $\bf{1.22\pm0.04}$ & $\bf{0.56\pm0.04}$ & $\bf{0.58\pm0.03}$ & $\bf{1.84\pm0.12}$ & $\bf{0.57\pm0.06}$ \\
lsir & $0.09\pm0.02$ & $1.74\pm0.07$ & $0.03\pm0.01$ & $0.16\pm0.04$ & $4.28\pm0.40$ & $0.03\pm0.01$\\
rand.lsir & $ 0.28\pm0.03$ & $1.34\pm0.06$ & $0.46\pm0.05$ & $0.50\pm0.04$ & $2.23\pm0.18$ & $0.48\pm0.06$ \\
pca & $0.20\pm0.03$ & $1.50\pm0.06$ & $0.31\pm0.04$ & $0.35\pm0.03$ & $3.03\pm0.26$ & $0.31\pm0.04$\\
rand.pca & $0.20\pm 0.03$ & $1.50\pm 0.06$ & $0.32\pm 0.04$ & $0.36\pm0.03$ & $3.00\pm0.26$ & $0.31\pm0.04$\\
\hline
\end{tabular}
\caption{Estimates of the mean performance and its standard error based on the Latent Factor Regression model. 
The results are based on 20 replicate data sets of dimension $n=500$, $p=3000$.}
\label{fr:plarge}
\end{table}

\subsection{Real data}\label{real:data:model}
In this section we evaluate the predictive performance of kNN classifiers
after dimension reduction by the proposed supervised and unsupervised
approaches. Our evaluation metric is the \% classification accuracy. In all data
analyses we set the number of slices for SIR and LSIR to be $H=10$ and assume 
that the dimension of the reduction subspace equals the rank of $\Gamma$.

\subsubsection{Digit Recognition}\label{real:data:digits}
The first data set we consider contains 60,000 digital images of
handwritten grey-scale digits from MNIST. This data set has been extensively
studied in the past and has been found to contain non-linear structure.
Each image has dimensions $28 \times 28$, which we represent in a vectorized
form, ignoring the spatial dependencies among neighboring pixels.
Hence $p = 28 \times 28 =784$. After removing the pixels that have
constant values across all sample images, the dimension becomes $p=717$.
For more details regarding the data refer to the Appendix.
The dimension reduction methods we compare are \textit{pca}, 
\textit{sir}, \textit{lsir}, and \textit{rand.lsir}, and use as a baseline
reference Gaussian random projections (\textit{rand.proj}). The value for 
$t^*$ is estimated from the data, assuming fixed value for the rank of $\Gamma$, 
$d^{*}$ (see Section \ref{rankest_unsup}). 
%The classification accurcy is reported over the full range of values for $d^{*}$-- the number edrs.
We fix the neighborhood size to be 10 for both the accuracy evaluation and
the slice estimation for LSIR. Each training set consists equal number of randomly
selected images of each digit and each test set is constructed in identical fashion.

\begin{figure}[h!tb]
\begin{minipage}{.4\textwidth}
\centering
\includegraphics[scale=0.47]
{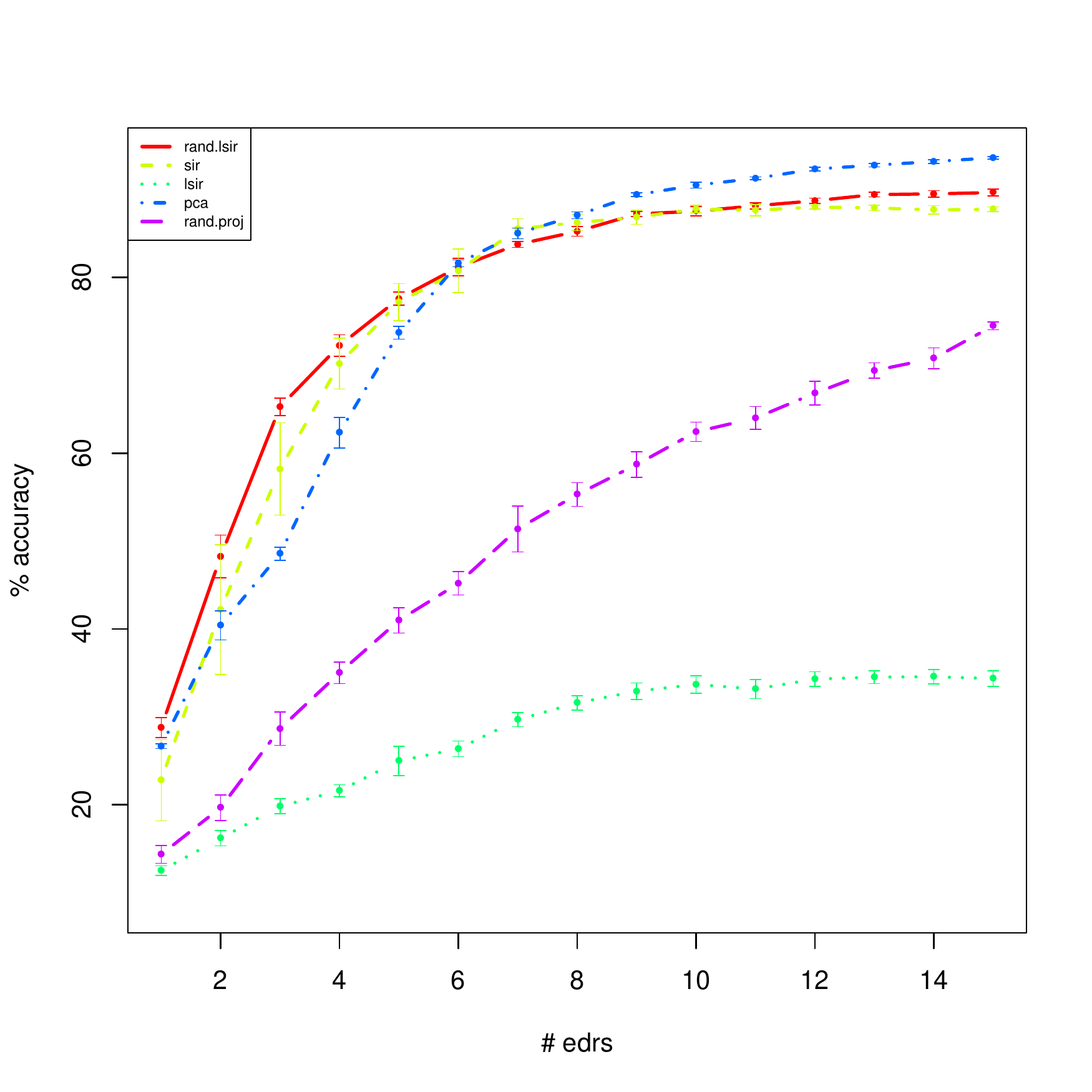}
\end{minipage}\qquad \qquad \qquad
\begin{minipage}{.4\textwidth}
\centering
\includegraphics[scale=0.47]
{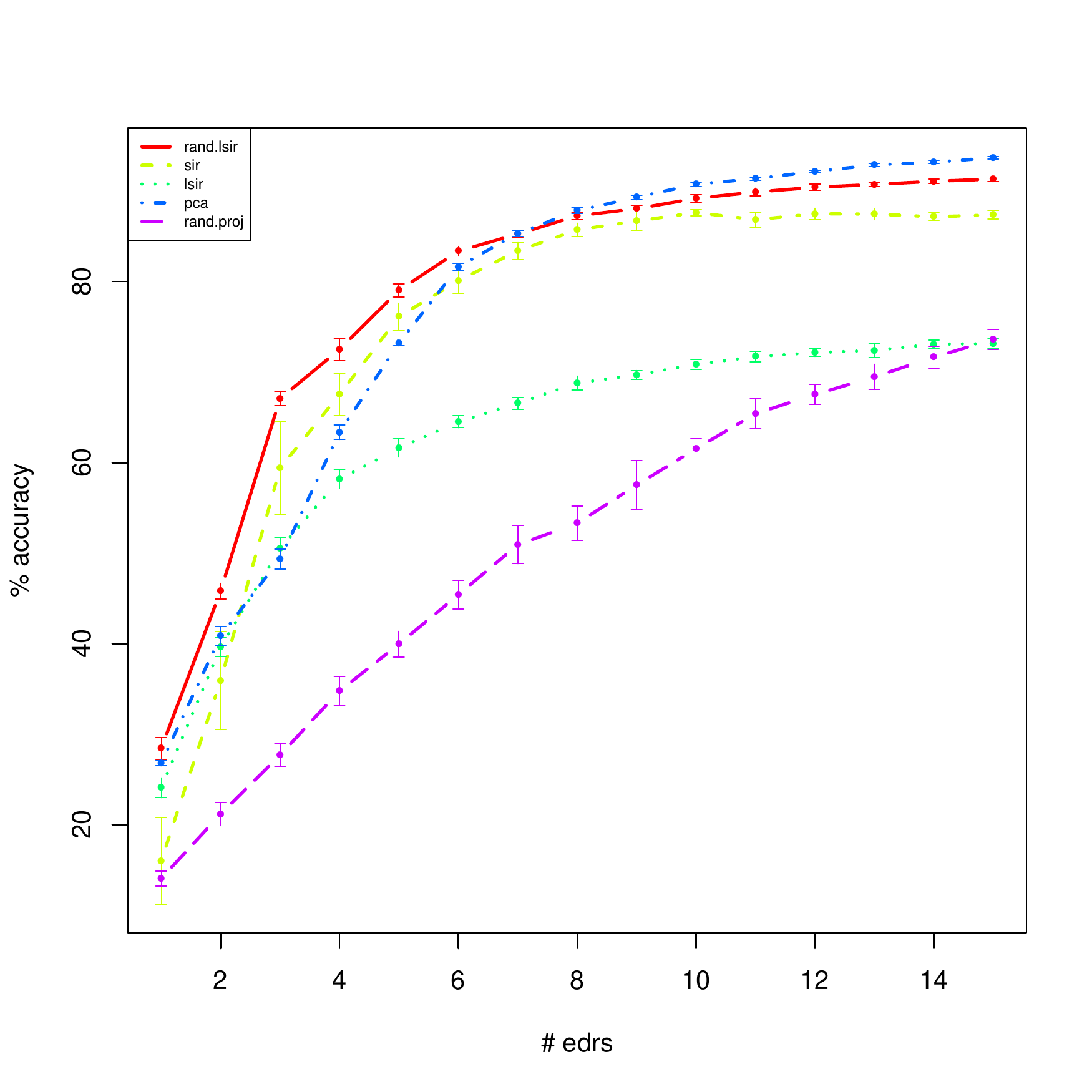}
\end{minipage}
\caption{Simulation results from 10 random data sets of size 50 (left panel) and 100 (right panel) randomly
selected samples for each digit. Hence the total sample size is $n=500$ and $n=1000$, respectively.
Each sample image is of dimension $p=717$. The error bars are based on 2 standard errors of the sample mean.}
\label{fig:digits}
\end{figure}

\paragraph{Results.}
The performance results for different ranks of $\Gamma$
are summarized in Figure \ref{fig:digits}. We use n=50 and n=100 training examples
for each of the digits between 0 and 9. Note that SIR can estimate up to 9 edrs 
(\# classes - 1), while LSIR-based approches do not have that limitation, so we 
report results allowing for the number of edrs to vary from 1 to 15 (fixing it at at a maximum of 
9  for SIR).
Randomized LSIR tends to perform the best for smaller number of edrs, with SIR and pca 
catching up as the number increases. Also, LSIR seems to be prone to overfitting and its 
performance improves with increased sample size. %We suspect this is again due
%to the implicit regularization due to the randomization.
%%%%%%%%
% Figure 5
%%%%%%%%

\subsubsection{HapMap gene expression data}
In this section we evaluate the classification accuracy of different
approaches on a microarray gene expression data set from
\citep{stranger:2012} for the purposes of e-QTL analysis,
associating genetic variants to variation in the gene expression.
Our focus is only on the gene expression data, using the
pre-processed values as features to classify the individual
samples to their respective population of origin. The data
set consists of gene expression data from lymphoblastoid
blood cells assayed in 728 individuals from 8
world populations (HapMap3): western Europe 112, Nigeria 108, 
China 80, Japan 82, India 81, Mexico 45, Maasai Kenya 137, 
Lihua Kenya 83. After pre-processing of the data for potential 
technical artifacts we use as input for the analysis 12,164 gene 
features. For more details regarding the pre-processing 
of the data refer to the Appendix \ref{app:hapmap}.
We set the neighborhood size to be 10 for both the
accuracy evaluation and the slice estimation for LSIR. 
The value of $t^*$ is estimated from the data 
(see Section \ref{rankest_unsup}) and the classification accuracy 
is reported over a range of values for the number edrs.

\begin{figure}[h!t!b]
\begin{center}
\includegraphics[scale=0.6]{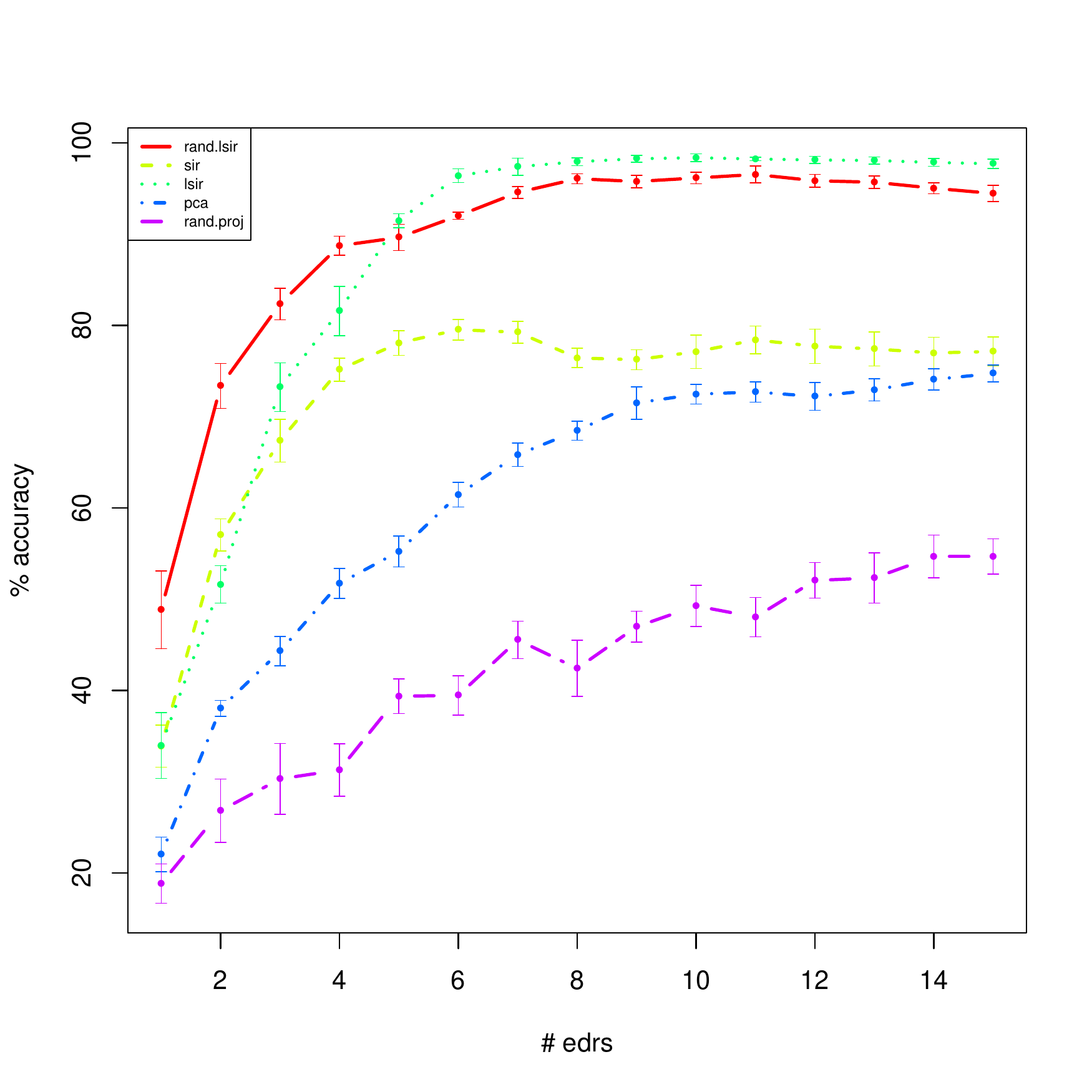}
\label{fig:xor_eq}
\end{center}
\caption{Classifiction accuracy of the k-nearest neighbor classifier applied to
microarray gene expression data from 8 HapMap populations stratified by population.
$n=728$, $p = 12,164$. The results are based on 10 random splits of the data into train
and test set of equal sizes. The x-axis contains the number of dimension reduction directions
used to project the data.}
\label{fig:hapmap}
\end{figure}

\paragraph{Results.}
Figure \ref{fig:hapmap} contains estimates of the estimation accuracy
for different number of dimension reduction directions. The estimates are
based on 10 random splits of the data into equally sized train and test set.
All approaches show large improvement upon the baseline random projection 
reference.
Clearly, rand.lsir and lsir outperform sir and pca, suggesting 
that there may be strong non-linear structure in the gene expression data, 
which is predictive of the population of origin. In additon, similarly to the 
digits data example from Section \ref{real:data:digits}, for smaller number 
of edrs the randomizated version of LSIR seems to produce the best 
classification performance, which becomes comparable to LSIR as the 
number of edrs increases. This suggests a potential advantage of randomized 
methods if a more parsimonious dimension reduction model, with few edrs,
is to be estimated.

%%%%%%%%
% Figure 6
%%%%%%%%
%\clearpage

\section{Discussion}
Massive high dimensional data sets are ubiquitous in modern applications.
In this paper we address the problem of reducing the dimensionality, in order
to extract the relevant information, when analyzing such data. The
main computational tool we use is based on recent randomized algorithms
developed by the numerical analysis community. Taking into account the presence of
noise in the data we provide an adaptive method for estimation of both the rank $d^*$
and number of Krylov iterations $t^*$ for the randomized approximate version of SVD.
Using this adaptive estimator of low-rank structure, we implement efficient algorithms for
PCA and two popular semi-parametric supervised dimension reduction methods -- SIR and LSIR.
Perhaps, the most interesting statistical observation is related to the implicit
regularization which randomization seems to impose on the resulting estimates.
Some important open questions still remain:
\begin{enumerate}
\item[(1)] There is need for a theoretical framework to quantify what generalization
guarantees the randomization algorithm can provide on out-of-sample
data and the dependence of this bound on the noise and the structure in
the data on one hand and on the parameters of the algorithms on the other;
\item[(2)] A probabilistic interpretation of the algorithm could contribute additional insights
into the practical utility of the proposed approach under different assumptions. In particular
it would be interesting to relate our work to a Bayesian model with posterior modes that
correspond to the subspaces estimated by the randomized approach.
\end{enumerate}

\section*{Acknowledgements}
SM would like to acknowledge Lek-Heng Lim, Michael Mahoney, Qiang Wu, and Ankan Saha.
SM is pleased to acknowledge support from grants NIH (Systems
Biology): 5P50-GM081883, AFOSR: FA9550-10-1-0436, NSF CCF-1049290,
and NSF-DMS-1209155.
SG would like to acknowledge Uwe Ohler, Jonathan Pritchard and Ankan Saha.

%\clearpage
\section*{Appendix}
\subsection*{Randomized algorithms for dimension reduction}
\floatname{algorithm}{Algorithm}
\vspace{5mm}
\begin{algorithm*}
\caption{: Randomized (L)SIR} \vspace{2mm}
\label{alg:RSIR_LSIR}
\vspace{2mm}
\textbf{input:}
$X \in \RR^{n \times p}$: data matrix\\
$H$: number of slices\\
$k$: number of nearest neighbors [LSIR only]\\
$d_{\max}$: upper bound for the dimension of the projection subspace (can provide a fixed value instead: $\hat{d}^{*}\equiv d^*$) \\
$\Delta$: oversampling parameter for the \textit{Adaptive Randomized SVD} step \ \ [default: $\Delta=10$]\\
$t_{\max}$: upper bound for the number of power iterations for the \textit{Adaptive Randomized SVD} step (can provide a fixed value instead) \\
\textbf{output:}
$\hat{G}_{(L)SIR} \in \RR^{p \times \hat{d}^{*}}$: a basis matrix for the effective
dimension reduction subspace\\
\\
\textbf{Stage 1}: Estimate low-rank approximation to $\hat{\Gamma}_{(L)SIR}$
\begin{enumerate}
\item Construct \ \ $J_{(L)SIR}$, as described in Section \ref{sec:exact:sdr}
\item Set \ \ $L=X^TJ_{(L)SIR}^T \ \ (\Rightarrow \hat{\Gamma}_{(L)SIR} = L L^T)$
\item Factorize \& estimate rank \ \ \{$[U,S,V]$, $\hat{d}^{*}$\} = $\textit{Adaptive Randomized SVD}(L, t_{\max}, d_{\max}, \Delta)$
\end{enumerate}
\textbf{Stage 2}: Solve the generalized eigendecomposition
\begin{enumerate}
\item Construct \ \ $A=SU^TX^T \in \RR^{\hat{d}^{*} \times p}$
\item Factorize \ \ $[\tilde{U},\tilde{S},\tilde{V}]=\text{full SVD(A)}$
\item Back-transform \ \ $\hat{G}_{(L)SIR} =\tilde{U} S U^T \in \RR^{p \times \hat{d}^{*}}$
\end{enumerate}
\end{algorithm*}
%\clearpage

\subsection*{XOR model}\label{app:xor}
Figure \ref{fig:xor_eq} includes an example of n=100 simulated data points
from the XOR example described in Section \ref{simulation:model:xor}.
The number of signal directions $d^*=4$. We show the projected data onto
the first two coordinate axes.

%%%%%%%%
% Figure 3
%%%%%%%%
\begin{figure}[h!t!b]
\begin{center}
\includegraphics[scale=0.75]{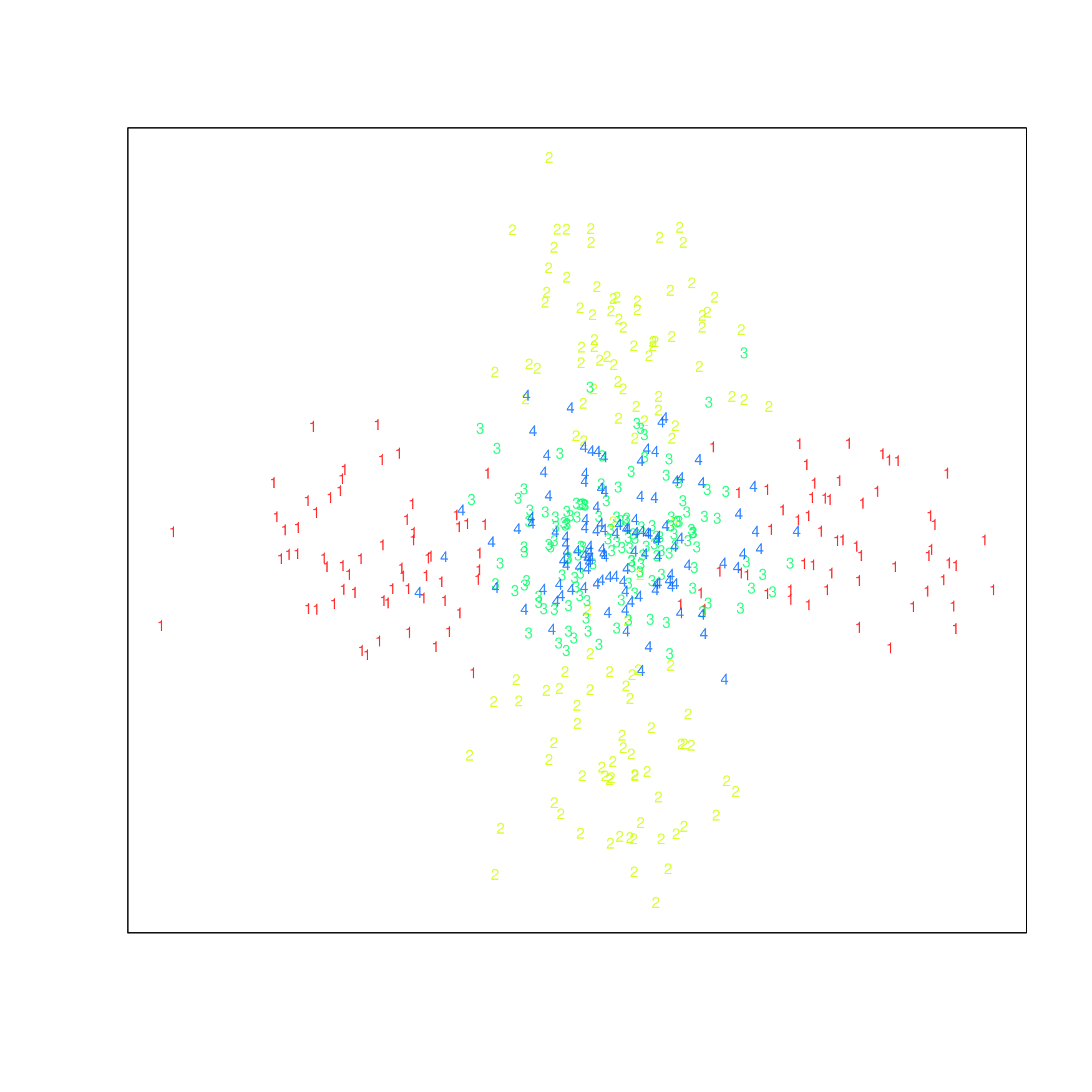}
\caption{XOR classification example. $n=100, d^*=4$. The data is projected onto the first two
coordinate directions of $X$.}
\label{fig:xor_eq}
\end{center}
\end{figure}
\subsection*{Digit Recognition}\label{app:digits}
The MNIST data set (Y. LeCun, htpp://yann.lecun.com/exdb/mnist/) \citep{lecun-98}
contains 60,000 images of handwritten grey-scale digits of dimension
$p=28 \times 28 = 784$. It is a subset of a larger set available from NIST.
The digits have been size-normalized and centered in a fixed-size image.
In our evaluations we use the subset of 'training images' which is contained
in files 'train-images-idx3-ubyte.gz' 'train-labels-idx1-ubyte.gz'.
\subsection*{HapMap microarray gene expression data}\label{app:hapmap}
The HapMap microarray gene expression data from from \citep{stranger:2012}
contains 2 replicates per individual, measured using the Illumina's Sentrix
Human-6 Expression BeadChip version 2 whole genome expression array
($\sim$ 48,000 probesets). We use as input to the pre-processing pipeline
the background-corrected and summarized (by the Illumina software)
probeset intensities, which map to unique genes and is annotated
as 'good' by the thorough re-annotation effort reported in
\citep{Lynch_and_Tavare2010} (using ENSEMBL gene annotation, release
GRCh37/hg19). This corresponds to 12,164 Illumina probeset expression values.
%In addition, we filter out probesets annotated to contain a SNP
%according to the re-annotation of the Illumina beadArrays platform described
%in \citep{Lynch_and_Tavare2010}, The output from the gene expression
%pre-processing is a set of 5,646 Illumina probeset expression values
%(5,323 genes).
\subsubsection*{Pre-processing pipeline}
This section describes the pre-processing steps for the gene expression data.
\begin{enumerate}
\item Impute non-positive probeset entries using the median intensity across all probeset within the same
population, independently for the individuals from Rep1 nd Rep2.
\item $log_2$-transform all probest measurements.
\item Select probesets marked as 'good' by the re-annotation of the Illumina
microarray data in \citep{Lynch_and_Tavare2010}.
\item Remove the top variance direction of the difference between technical replicates:
Rep1-Rep2 from Rep1 and from Rep2 -- results in large improvement in the
correlation between Rep1 and Rep2 across all samples
%\item Center the distribution of probeset intensities for all individuals within each population to zero mean.
%\item Normalize probeset values across individuals: Quantile-normalize, independently for Rep1 and Rep2, the
%probeset values across all arrays to be distributed as N(0,1).
\item Average probeset values between Rep1 and Rep2.
%\item Regress out the top 2 gene expression sample PCs, to account for potential remaining
%technical artifacts in the data.
\end{enumerate}

\clearpage

\subsection*{Unsupervised dimension reduction}
In this section we outline a potential extension of the ideas for randomized
dimension reduction developed in the main text to the unsupervised problem
of manifold learning. In particular, we focus on Localized Locality Projections
\citep{LPP:NIPS:2003}, which is a linear relaxation of the Laplacian Eigenmaps
method originally introduced in \citep{misha1} and allows for easy
out-of sample application of the estimated reductions.
\paragraph{Dimension reduction based on graph embeddings (LPP)}
Dimension reduction based on graph embeddings seek to map the original data points
to a lower dimensional set of points while preserving neighborhood relationships.
The theoretical assumptions underlying these methods are that the data lies on a
smooth manifold embedded in the high-dimensional ambient space. This manifold is unknown
and needs to be inferred from the data. Given sufficient number of observations the manifold can
be reasonably represented as a $G=(E,V)$ \citep{Chung_1997} where the vertexes
$\{v_1,..,v_n\}$ correspond to the $n$ observations $\{x_1,..,x_n\}$ and the edges
corresponds to points which are close to each other. For example, this neighborhood
relationship can be encoded in a sparse symmetric adjacency matrix $W$. Given $W$, the
Laplacian Eigenmaps (LE) algorithm \citep{misha1} embeds the data into a low dimensional
space preserving local relationships between points.
Given the adjacency or association matrix the Graph Laplacian is constructed
$L=D-W$, where $D$ is a diagonal matrix with $D_{ii} = \sum_j W_{ij}$.
A spectral decomposition of $L$
\begin{equation}
\label{le:asm}
L v_i = \lambda_i v_i
\end{equation}
results in eigenvalues $\lambda_1 =0 \leq \lambda_2 \leq ... \leq \lambda_n$
with $v_1 = \mathbf{1}$. Projecting the matrix $L$ onto the $d^*$ eigenvectors
corresponding to the smallest $d^*$ eigenvalues greater than zero embeds
the $n$ points into a $d^*$ dimensional space. Notice that (\ref{le:asm}) is
a special case of (\ref{eqn:gsvd:exact1}), with $\Gamma = L$ and $\Sigma=I$.
Under certain conditions
\citep{Belkin:Niyogi:2008}, the Graph Laplacian converges to the Laplace-Beltrami
operator on the underlying manifold. This provides a theoretical
motivation for the embedding. This embedding needs to be recomputed when
a new data point is introduced and typically will not be a linear projection of the
data. Hence, for computational reasons it would be advantageous to have a linear projection
that can be applied to new data points without having to recompute the spectral
decomposition of the graph Laplacian. The goal of Locality Preserving Projections
\citep{LPP:NIPS:2003} is to provide such a linear approximation to the non-linear
embedding of Laplacian Eigenmaps \citep{misha1}.
The dimension reduction procedure starts by specifying the dimension of the
transformed space to be $d^* < n$. Let the parameter defining the neighborhood size
be $k$. Locality Preserving Projections (LPP) \citep{LPP:NIPS:2003}
is stated as the following generalized eigendecomposition problem.
\begin{align}
\label{eqn:lpp}
\begin{split}
X^TLX e &= \lambda X^TDX e, \\
X^T(D-W)X e	&= \lambda X^TDX e, \\
X^TWX e	 &= (1-\lambda) X^TDX e, \\
\tilde{X}^T\tilde{W}\tilde{X} e &= \mu \tilde{X}^T\tilde{X} e,
\end{split}
\end{align}
where, $\tilde{X}=D^{\frac{1}{2}}X$,
$\tilde{W} = D^{-\frac{1}{2}}WD^{-\frac{1}{2}}$, $\mu = 1 - \lambda$ and
for a fixed bandwidth parameter $b > 0$,
$$W_{ij} = \left\{
\begin{array}{l l}
\text{exp}\left\{-\frac{||X_{i} - X_{j}||} {\text{b}}\right\} & \quad
\text{if $X_i$ is among the $k$-NN of $X_j$ or vice versa}\\
0 & \quad \text{otherwise}.\\
\end{array} \right.
$$
The column vectors that are the solutions to equation \eqref{eqn:lpp} (excluding the trivial solution $e_0$)
are the required embedding directions $\{e_j\}$, ordered according to their
generalized eigenvalues $1=\mu_0 > \mu_1 > \ldots > \mu_{d^{*}-1}$.
Hence the neighborhood-preserving optimal embedding according to the LPP criterion is:
$$x_i \rightarrow A^Ty_i,\ \ \ \text{where} \ \ A=(e_1,\ldots,e_{d^{*}}).$$
The formulation (\ref{eqn:lpp}) is a special case of (\ref{eqn:gsvd:exact1}), with
$\Gamma=\tilde{X}^T\tilde{W}\tilde{X}$ and $\Sigma = \tilde{X}^T\tilde{X}$.

\paragraph{LPP Estimation.}
The approximation algorithm to solve LPP requires solving the generalized
eigendecomposition stated by the last equation in derivation (\ref{eqn:lpp}).
There will be on the order of $kn$ non-zero entries. Hence, the matrix $\tilde{W}$
will be sparse, assuming the size of the local neighborhoods $k \ll n$. This implies that
the matrix product $X^TWX \Omega$, where $\Omega \in \RR^{p \times l}$ can be
efficiently computed in $O((k+l)np)$ time, without explicitly constructing $W$
by using the $k$-nearest neighbor matrix. Hence $\tilde{\Gamma}:=\tilde{X}^T\tilde{W}\tilde{X}$
can be efficiently approximated using \textit{Adaptive Randomized SVD} algorithm
from Section \ref{sec:rsvd} which would provide and estimate for $SU^T$. Then,
setting $\tilde{\Sigma}:=\tilde{X}^T\tilde{X}$, we reduce the problem to 
(\ref{eqn:symm:gsvd}). Next we describe the randomized LPP algorithm 
in more detail:

\floatname{algorithm}{Algorithm }
\vspace{5mm}
\begin{algorithm*}
\caption{: Randomized LPP} \vspace{2mm}
\label{alg:RLPP}
\vspace{2mm}
\textbf{input:}
$X \in \RR^{n \times p}$: data matrix\\
$k$: number of nearest neighbors capturing local manifold structure\\
$d_{\max}$: upper bound for the dimension of the projection subspace (can provide a fixed value instead) \\
$\Delta$: oversampling parameter for the \textit{Adaptive Randomized SVD} step \ \ [default: $\Delta=10$]\\
$t_{\max}$: upper bound for the number of power iterations for the \textit{Adaptive Randomized SVD} step (can provide a fixed value instead) \\
\textbf{output:}
$\hat{G}_{LPP} \in \RR^{p \times d^{*}}$: a basis matrix for the embedding subspace\\
\\
\textbf{Stage 1}: Estimate low-rank approximation to $\tilde{X}^T \tilde{W} \tilde{X}$
\begin{enumerate}
\item Construct $\tilde{X}$ and $\tilde{W}$,
\item Factorize \{$[U,S,U]$, $\hat{d}^{*}$\} = $\textit{Adaptive Randomized SVD}(\tilde{X}^T \tilde{W} \tilde{X}, t_{\max}, d_{\max}, \Delta)$
\end{enumerate}
\textbf{Stage 2}: Solve the generalized eigendecomposition
\begin{enumerate}
\item Construct $A=SU^T\tilde{X}^T \in \RR^{\hat{d}^{*} \times p}$
\item Factorize $[\tilde{U},\tilde{S},\tilde{V}]=\text{full SVD(A)}$
\item Back-transform \ \ $\hat{G}_{LPP} =\tilde{U} S U^T \in \RR^{p \times \hat{d}^{*}}$
\end{enumerate}
\end{algorithm*}

\clearpage

\bibliographystyle{chicago}
\bibliography{randalg}
\end{document}